\documentclass[journal]{packages/IEEEtran}
\usepackage{algorithm,algpseudocode}

\usepackage[numbers,sort]{natbib}
\usepackage{multicol}
\usepackage[bookmarks=true]{hyperref}
\usepackage[table]{xcolor} \usepackage[section]{placeins} 

\usepackage{graphics} \usepackage{amsmath} \usepackage{amssymb}  \usepackage{makeidx} \usepackage{graphicx} \graphicspath{{figures/}}
\usepackage{multicol} \usepackage[utf8]{inputenc} \usepackage{multirow}
\usepackage{booktabs} \usepackage{adjustbox}
\usepackage{subcaption}
\usepackage{bm} \usepackage{color} \usepackage{empheq} \usepackage{siunitx} 

\usepackage{blindtext}

\usepackage{todonotes} 

\usepackage{paralist}

\usepackage{mathtools}
\DeclarePairedDelimiterX{\norm}[1]{\lVert}{\rVert}{#1}

\title{Visual-Inertial SLAM as Simple as A, B, VINS}

\author{Nathaniel Merrill and Guoquan Huang\thanks{The authors are with the Robot Perception and Navigation Group (RPNG), University of Delaware, Newark, DE 19716, USA. Email: \tt{\{nmerrill, ghuang\}@udel.edu}}}

\begin{document}

\maketitle

\begin{abstract}
We present AB-VINS, a different kind of visual-inertial SLAM system.
Unlike most popular VINS methods which only use hand-crafted techniques, AB-VINS makes use of three different deep neural networks.
Instead of estimating sparse feature positions,
AB-VINS only estimates the scale and bias parameters ($\mathbf a$ and $\mathbf b$) of monocular depth maps, as well as other terms to correct the depth using multi-view information,  which results in a compressed feature state.
Despite being an optimization-based system, the front-end motion tracking thread of AB-VINS surpasses the efficiency of a state-of-the-art filtering-based method while also providing dense depth. 
When performing loop closures, standard keyframe-based SLAM systems need to relinearize a number of variables which is linear with respect to the number of keyframes.
In contrast, the proposed AB-VINS can incorporate loop closures while only affecting a constant number of variables.
This is thanks to a novel data structure called  the memory tree, where keyframe poses are defined relative to each other rather than all in one global frame, allowing for all but a few states to be fixed.
While AB-VINS might not be as accurate as state-of-the-art VINS algorithms, it is shown to be more robust.
\end{abstract}

\section{Introduction} \label{sec:intro}

Visual-inertial SLAM (or VINS) is essential for autonomous robots, augmented reality (AR) and virtual reality (VR) or extended reality (XR)~\cite{Huang2019ICRA}.
Leveraging the complimentary camera and IMU measurements, such systems output high-rate, metric, gravity-aligned poses (orientations and positions) required for XR rendering, while bounding the pose estimation error over time by performing loop closures to past states in order to prevent the user from drifting away in the virtual world.
Besides the poses, dense depth is useful, which, for instance, helps to create the illusion of occlusion with rendered object and  solid surfaces for virtual characters to interact with.
In  robotic applications, poses paired with dense depth are also needed, for example, for obstacle avoidance and path planning.

Despite the need for dense depth, few VINS solutions are able to provide it.
Additionally, despite recent advances in deep learning for a multitude of related applications~\cite{Ranftl2022TPAMI,Arandjelovic2016CVPR,Detone18ArXiV}, very few VINS methods have utilized it.
AB-VINS on the other hand provides high-rate gravity-aligned poses along with dense depth, and uses three different deep networks to its advantage.
Visual-inertial SLAM often needs to be as efficient as possible, being able to quickly close loops while running on some of the smallest devices -- even as small as a bee robot in future applications.
The proposed AB-VINS contains, to the best of our knowledge, the most efficient pose graph SLAM solution thanks to the novel memory tree data structure, which brings us one step closer to achieving such goals.

\begin{figure}
\centering
\includegraphics[width=\columnwidth,trim={0 0 0.1cm 0},clip]{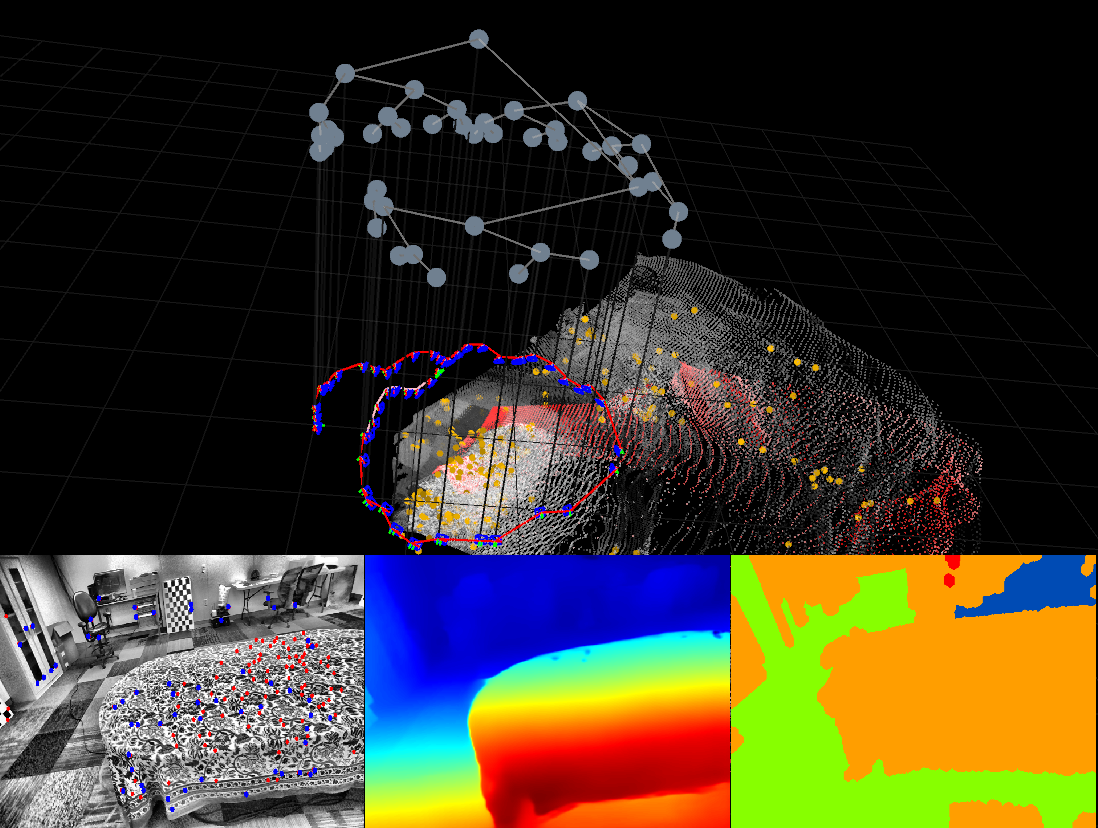}
\caption{
Visualizing the proposed AB-VINS:
On the bottom left is the current frame with feature tracks (red) and reprojected points (blue) overlayed.
On the bottom center is the most recent keyframe depth map.
On the bottom right is the code mask, which divides the images into different regions that are pushed and pulled to correct the depth according to multi-view information.
In the main window at the top the memory tree can be seen, which is a novel data structure used to speed up pose graph optimization.
}
\label{fig:overview}
\end{figure}

In particular, Fig.~\ref{fig:overview} depicts the overall system of the proposed AB-VINS which has the following main contributions:
\begin{itemize}
    \item A new monocular visual-inertial SLAM system AB-VINS is proposed, which exhibits state-of-the-art robustness and efficiency while also providing dense depth.
    \item A compact feature representation called AB features is presented, where an arbitrary number of feature positions are parameterized by $a$ and $b$, the scale and bias of a monocular depth map, as well as other optional terms to correct the depth with multi-view information.
    \item The memory tree, a novel data structure used to significantly speed up pose graph optimization, is introduced. The memory tree allows AB-VINS to solve pose graph SLAM while only relinearizing a {\em constant} number of variables, which is proven experimentally.
\end{itemize}

The rest of the paper is organized as follows: 
After reviewing the related work in Sec.~\ref{sec:relworks}, we present in detail the proposed AB-VINS  in Sec.~\ref{sec:method} and \ref{method2}.
In Sec.~\ref{sec:experiments} AB-VINS is thoroughly evaluated for accuracy, efficiency, and robustness.
Finally, in Sec.~\ref{sec:conclusion}, closing remarks are made.

\section{Related Work} \label{sec:relworks}

In this section, we provide a complete review of the literature that is closely related to 
the proposed AB-VINS, including visual-inertial SLAM, dense SLAM and efficient pose graph optimization methods.

\subsection{Visual-Inertial SLAM}

A multitude of visual-inertial SLAM systems have been developed in recent years~\cite{Huang2019ICRA}.
The most similar one in nature to our AB-VINS is VI-ORB-SLAM~\cite{Mur2017RAL}, 
which is incorporated into ORB-SLAM3~\cite{Campos2021TRO}.
Similar to AB-VINS, VI-ORB-SLAM is an optimization-based system and has three modules: tracking, local mapping, and mapping.
Despite using one more CPU thread than AB-VINS, it does not provide dense depth (and actually none of the systems mentioned below do so).
VINS-Mono~\cite{Qin2018TRO}, and later VINS-Fusion\footnote{Available at: \url{https://github.com/HKUST-Aerial-Robotics/VINS-Fusion}},
similarly to AB-VINS have two threads, one for visual-inertial odometry (VIO) and one for mapping, although VIO can be performed on two threads if desired.
Another visual-inertial SLAM system, ROVIOLI~\cite{Schneider2018RAL}, also has a VIO front-end and pose graph back-end, and supports multi-session mapping and localization.
The recent work~\cite{Huai2024TRO} similarly has a VIO front-end and optimization-based back-end, and aims at producing state estimates with covariances that are consistent with the true estimation error.
BASALT~\cite{Usenko2020RAL} attempts to solve the problem of degraded IMU information for keyframes, and solves loop closures in a global bundle adjustment (BA), which may be computationally expensive.
OKVIS2~\cite{Leutenegger2022ArXiV}  utilizes a semantic segmentation network to ignore features that should not be tracked (such as on clouds).
The main drawback of all of these methods is that the number of variables affected by the global optimization grows linearly with the number of keyframes, while AB-VINS can solve loop closures while only relinearizing a {\em constant} number of variables.

Unlike visual-inertial SLAM, pure VIO methods do not have a memory about places, and thus the system steadily drifts over time -- but is highly accurate locally.
Many VIO systems have been proposed~\cite{Huang2019ICRA}.
The multi-state constraint Kalman filter (MSCKF)~\cite{Mourikis2007ICRA} is one of the first such examples.
MSCKF avoids estimating the feature positions in the EKF framework, which can be expensive with more than just a few features, by applying the nullspace projection to eliminate the dependency on the feature states.
Over the years, MSCKF has been adapted and improved, for example, by adding small amounts of structure to the state~\cite{Li2012IROS}, adding online calibration and improving consistency~\cite{Li2013IJRR}, and utilizing the efficient sparse square root inverse covariance filter~\cite{Wu2015RSS}.
Robocentric VIO (R-VIO)~\cite{Huai2019IJRR}, which is also based on the EKF, showed that estimating the poses in a local robocentric frame (the oldest frame in the sliding window) rather than in the global as usual has better observability properties.
R-VIO partially inspired the anchored local mapping window of AB-VINS (discussed in Sec.~\ref{sec:method:local_mapping:optimization}), however instead of defining the anchored frame in the last frame of the sliding window, we define it in the gravity-aligned frame centered at the oldest frame in the window so that the local gravity direction does not have to be estimated.
Another popular type of VIO system is based on nonlinear optimization,
which often marginalizes states that are removed from the optimization in order to maintain a large covariance matrix for additional prior cost terms~\cite{Leutenegger2014IJRR, Leutenegger2022ArXiV, Qin2018TRO, Usenko2020RAL}.
However, the computational cost of marginalization is high in optimization-based systems compared to EKF if calculating the marginal covariance~\cite{Chen2024ICRA}, 
motivating us to choose to ommit marginalization in AB-VINS and instead opt to simply fix the oldest pose in the sliding window similarly to ORB-SLAM.

\subsection{Dense SLAM}

Dense VINS can provide both high-rate gravity-aligned poses and dense depth.
Kimera~\cite{Rosinol2020ICRA} provides dense depth via Delaunay triangulation on the sparse VIO points, however this is an oversimplification of the scene and can not capture small details of the structure.
CodeVIO~\cite{Zuo2021ICRA} is a dense VIO system which, similarly to AB-VINS, estimates correction terms for the depth within the state, but the correction terms (codes) are learned from an autoencoder network.
The codes in AB-VINS are more hand-crafted, but can more easily push or pull entire objects with only a few feature measurements.
DiT-SLAM~\cite{Zhao2022Sensors} showed that the code dimension can be as low as 8.
In AB-VINS, on the other hand, the code dimension is 4.
SimpleMapping~\cite{Xin2023ISMAR} utilizes a deep multi-view stereo (MVS) network to provide dense depth, but does not utilize monocular priors to help the system when parallax is not available.

Pure visual dense SLAM systems have also been proposed.
The scale can not be estimated using only classical methods for monocular visual SLAM.
DTAM~\cite{Newcombe2011ICCV} is the first example of such a system, which estimates the pose via dense image alignment, and creates dense depth maps with classical MVS.
TANDEM~\cite{Koestler2021CoRL}, which pairs an MVS network with monocular visual odometry, and receives localization measurements from the TSDF map it maintains from fusing the dense depth.
Some visual SLAM systems, like CNN-SLAM~\cite{Tateno2017CVPR}, BA-Net~\cite{Tang2019ArXiV}, CodeSLAM~\cite{Bloesch2018CVPR}, and Deep Factors~\cite{Czarnowski2020RAL} can estimate metric scale by training a monocular depth network with metric depth supervision.
However, training a network in such a way does not benefit from the robustness that eliminating the scale and bias from monocular depth as in MiDaS~\cite{Ranftl2022TPAMI} provides.
AB-VINS can estimate metric scale because of the IMU, and still benefits from the robustness of MiDaS.

\begin{figure*}[t]
\centering
\includegraphics[width=\textwidth,trim={0.5cm 0.5cm 2.25cm 0cm},clip]{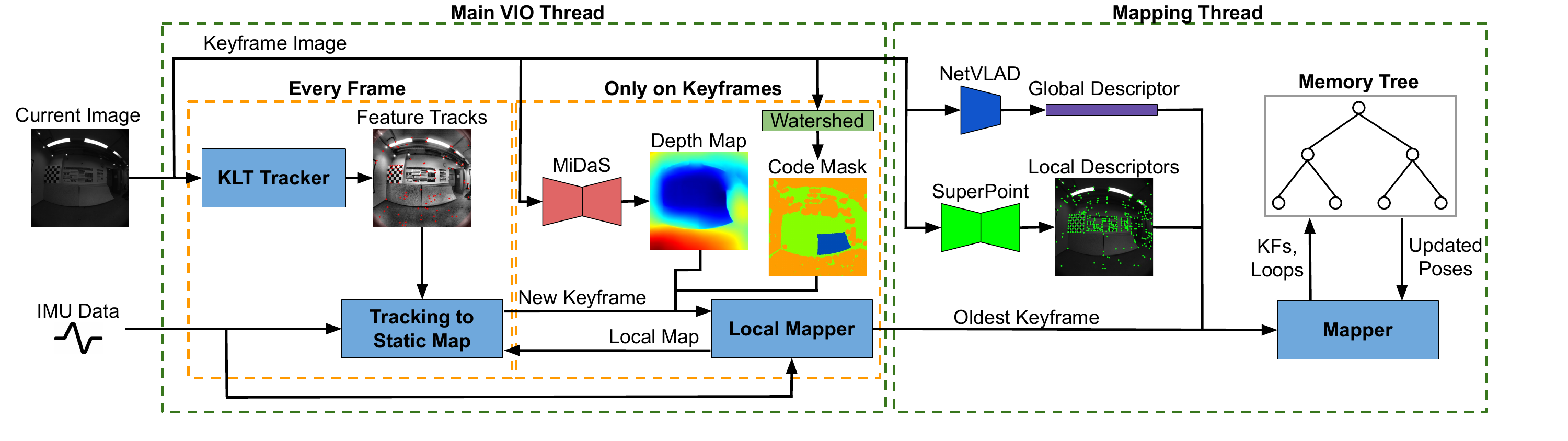}
\caption{System diagram of the proposed AB-VINS.}
\label{fig:sys}
\end{figure*}

\subsection{Efficient Pose Graph Optimization}

One of the first efficient solutions to pose graph optimization is the sparse pose adjustment (SPA)~\cite{Konolige2010IROS}, which utilizes the sparse information matrix structure in a Levenberg–Marquardt (LM) optimization to its advantage.
Another option is to utilize the square-root information form as in iSAM~\cite{Kaess2008TRO} and iSAM2~\cite{Kaess2011IJRR}.
iSAM2 introduced a data structure called  Bayes tree, which is used to reduce the number of times variables need to be relinearized by using the tree to identify which parts of the graph are affected by the measurements -- thus making optimization highly-efficient.
The proposed memory tree is not related to the Bayes tree since the keyframe poses are defined in a relative frame rather than the global.
SPA, iSAM, and iSAM2, as well as most other factor graph optimization methods ever proposed, define all of the poses in the global frame, which requires at least adjusting all of the poses in-between two loop keyframes.
Most similar to the proposed memory tree data structure is the relative method~\cite{Sibley2010RSS}, where a keyframe's pose is defined relative to the previous one's frame of reference.
While the relative method~\cite{Sibley2010RSS} was only applied to an incremental BA and not pose graphs, it could theoretically be used for pose graphs as well -- just like the memory tree could be applied to an incremental BA.
The relative method allows for the majority of the poses to be fixed, but the proposed memory tree has better algorithmic properties such as theoretically smaller maximum state size and time to compute the global pose as well as optimization with better worst-case complexity.
 \section{Tracking and Local Mapping} \label{sec:method}

We now explain the proposed AB-VINS  in detail.
Overall, as shown in Fig.~\ref{fig:sys}, the system is composed of two threads: VIO and mapping.
For every frame, the main VIO thread tracks the pose to the current static local map.
When a frame is selected to be a keyframe, a depth map and code mask are generated and then sent to the anchored local mapping module, which optimizes the local structure and pose with both visual and inertial measurements in a sliding window fashion along with online calibration.
Once a keyframe is popped off of the local mapping window, it is combined with the outputs of two more neural networks (i.e., one for global descriptors and the other for local keypoint descriptors) before being sent to the mapper, which communicates with the proposed memory tree and is responsible for finding loop closures.
In the following, we will first focus on the main VIO thread, while the mapper thread will be discussed in the next section.

\subsection{Fast Initialization}

While some recent methods have utilized learned monocular depth to improve visual-inertial initialization~\cite{Merrill2023RSS,Merrill2024IJRR,Zhou2022ECCV}, they still wait some time to initialize and require at least a small amount of motion.
In AB-VINS, we adopt a much simpler initializer by assuming the IMU biases are known well enough to integrate (which can be obtained offline by letting the device sit still).
We first estimate the gravity direction by running Gram–Schmidt on the first few accelerometer measurements.
While this does make a static assumption, we found that it works well except in the most dynamic scenarios, which may not be typical for AR/VR headsets or mobile robotic applications (e.g., indoor wheeled robots).
Then, we initialize the gravity-aligned map with the first frame's monocular depth using an arbitrary scale and bias (i.e., $a_0$ and $b_0$).
Finally, the velocity is set to zero, and tracking to the static map commences.
Thus, we can typically initialize using just the first frame.
No motion is required to initialize, but some excitation is needed to estimate the metric scale.
However, we argue that with very little motion (sitting mostly or completely still) scale is not as important as it is with large motions.

\subsection{Tracking to Static Map}

\begin{figure}
\centering
\includegraphics[width=0.8\columnwidth,trim={20.5cm 2cm 16cm 5cm},clip]{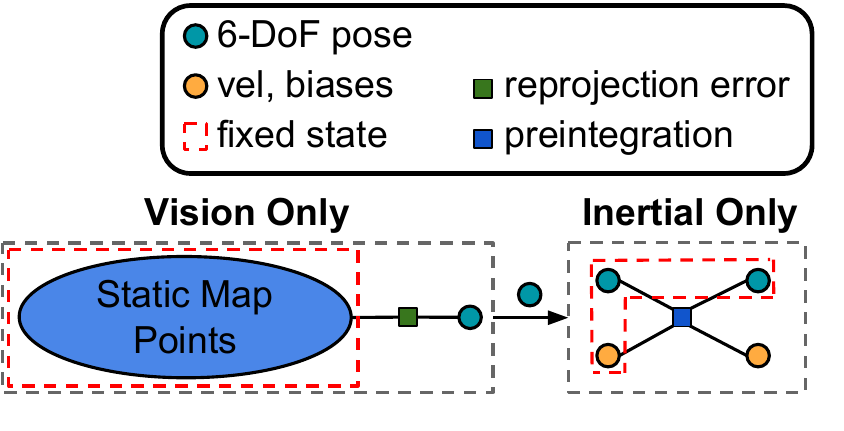}
\caption{A graphical representation of tracking to static map. The 6-DoF pose is estimated from the static map using only visual measurements, and an inertial-only optimization estimates the velocity and biases.}
\label{fig:static_map_tracking}
\end{figure}

For an incoming frame at time $t_k$, feature tracks are first obtained using the efficient Kenade-Lucas-Tomasi (KLT) tracking algorithm~\cite{Lucas1981IJCAI} on FAST corner points~\cite{Rosten2006ECCV}.
Then, the current pose is predicted from preintegrated measurements between the last frame and the current.
After that, tracking to the static local map occurs.
A visualization of this process can be seen in Fig.~\ref{fig:static_map_tracking}.

First, the 6-DoF pose, represented by the JPL quaternion~\cite{Trawny2005_Q_TR} ${}_A^{I_k}\bar{q}$ and 3D position vector ${}^A\mathbf{p}_{I_k}$, are estimated with only visual measurements by reprojecting points from the static local map.\footnote{Note that since the local map is scaled and gravity-aligned, the pose output of the vision-only optimization is scaled and gravity-aligned since it is equivalent to the PnP problem with a good initial guess from the IMU.}
Here, $\{A\}$ is the gravity-aligned anchor frame of the local mapping window and $\{I_k\}$ is the IMU frame at time $t_k$.
Formally, the vision-only optimization attempts to minimize a cost function $\mathcal{C}_{vo}$ given by:
\begin{align}
    \mathcal{C}_{vo} = \sum_{\ell \in \mathcal{V}_k} \lambda_\ell \rho_c\left(\norm{\mathbf{r}^{rs}_{\ell k}}^2_{\boldsymbol{\Sigma}_{v}}\right)
\end{align}
where $\mathcal{V}_k$ is the index set of all 3D points ${}^A\mathbf{p}_{f_\ell}$ tracked in the current frame, $\lambda_\ell$ is the edge weight proposed by~\cite{Zhou2022ECCV}, which down-weights the cost for residuals corresponding to depth points that are near the image edge or have a high depth edge value.
The edge images correspond to the depth map that the 3D point ${}^A\mathbf{p}_{f_\ell}$ originates from.
The function $\rho_c$ is the robust Cauchy loss function, 
and $\boldsymbol{\Sigma}_{v}$ is the diagonal noise covariance matrix for the residual
$\mathbf{r}^{rs}_{\ell k}$ which is defined as:
\begin{align}
    \mathbf{r}^{rs}_{\ell k} = \mathbf{u}_{\ell k} - h_{rs}({}_A^{I_k}\hat{\bar{q}}, {}^A\hat{\mathbf{p}}_{I_k}, {}^A\mathbf{p}_{f_\ell}, {}^C_I\bar{q}, {}^C\mathbf{p}_{I}, \bm{\zeta})
\end{align}
where $h_{rs}$ projects the point ${}^A\mathbf{p}_{f_\ell}$ into the current frame using the current pose estimate, $\mathbf{u}_{\ell k}$ is the raw pixel coordinate of the current feature track location for the $\ell$-th 3D point, and ${}^C_I\bar{q}, {}^C\mathbf{p}_{I}, \bm{\zeta}$ are the camera extrinsic and intrinsic calibration parameters from the local mapper.
Note that the 3D point and calibration are fixed for this optimization.
The optimization is run up to four times, and after each step each residual is run through a $\chi^2$ check.
If the residual fails the $\chi^2$ check, it is removed from the next round of optimization.
At most 50\% of residuals can be removed this way.
This process robustifies against bad feature tracks or noisy 3D points.

After the vision-only optimization, the velocity ${}^A\mathbf{v}_{I_k}$ and IMU biases $\mathbf{b}_{g,k}$, $\mathbf{b}_{a,k}$ are estimated in an inertial-only optimization (only preintegration measurements) with poses of the previous keyframe and current frame fixed.
For the preintegration, the angular velocity is integrated using the quaternion integrator~\cite{Trawny2005_Q_TR} and the acceleration is double integrated using fourth order Runge-Kutta.
In the case where the IMU measurements (angular velocity or linear acceleration) exceed a saturation threshold, the corresponding measurement noise is inflated to robustify the system against IMU saturation.
In particular, the inertial-only optimization tries to minimize the following cost function $\mathcal{C}_{io}$:
\begin{align}
    \mathcal{C}_{io} = \norm{\hat{\bar{\mathbf{x}}}_k \boxminus h_p(\mathbf{x}_i, {}^i\boldsymbol{\alpha}_k, {}^i\boldsymbol{\beta}_k)}^2_{\mathbf{Q}_{ik}}
\end{align}
which is the preintegration~\cite{Lupton2012TRO,Forster2015RSS,Eckenhoff2019IJRR} cost function between the previous keyframe at time $t_i$ and the current frame at  $t_k$.
We have the state $\hat{\bar{\mathbf{x}}}_k = \begin{bmatrix} {}_A^{I_k}{\bar{q}}^\top & {}^A{\mathbf{p}}_{I_k}^\top & {}^A\hat{\mathbf{v}}_{I_k}^\top & \hat{\mathbf{b}}_{g,k}^\top &\hat{\mathbf{b}}_{a,k}^\top \end{bmatrix}^\top$ which is an inertial state (note that the pose is fixed), and similarly for the completely fixed keyframe state $\mathbf{x}_i$.
The terms ${}^i\boldsymbol{\alpha}_k$ and ${}^i\boldsymbol{\beta}_k$ are the preintegrated measurements between $t_i$ and $t_k$ defined as~\cite{Eckenhoff2019IJRR}:
\begin{align}
{}^{i} \bm \alpha_{k} &=
\int_{t_i}^{t_{k}} \int_{t_i}^{s} {}^i_{u}\Delta\mathbf{R}\left(\mathbf{a}_m(u)- \mathbf{b}_a(u)-\mathbf{n}_a(u)\right) du ds \notag \\
{}^{i} \bm \beta_{k} &=
\int_{t_i}^{t_{k}} {}^i_{u}\Delta\mathbf{R}\left(\mathbf{a}_m(u)- \mathbf{b}_a(u) -\mathbf{n}_a(u)\right) du.
\end{align}
$\mathbf{Q}_{ik}$ is the covariance matrix for the preintegration residual, which can be inflated as discussed if  IMU saturation detected.

\begin{figure}
\centering
\includegraphics[width=0.45\columnwidth]{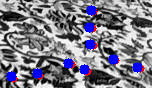}
\includegraphics[width=0.45\columnwidth]{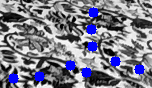}
\caption{\textbf{Left:} The result of tightly coupling the visual and inertial measurements in the tracking optimization. \textbf{Right:} The result of the proposed decoupled approach. Tracked image coordinates are in red and reprojected points are in blue.}
\label{fig:tight_vs_decoupled}
\end{figure}

Note that in the above we have chosen to optimize motion by decoupling the visual and inertial measurements.
This is because we found that tightly coupling the visual and inertial measurements in motion tracking (e.g., as in~\cite{Mur2017RAL}) can sometimes pull the solution away from one that satisfies the reprojection error, as seen in Fig.~\ref{fig:tight_vs_decoupled}, which creates wobblier pose output than desired for many applications.
The problem is exacerbated when tracking a small number of features as in our system -- since the vision terms can not outweigh the inertial -- and the noisiness of the monocular depth does not help either.
We emphasize that our decoupled approach is only possible with nonlinear optimization and not with filtering, as filters naturally tightly couple the measurements by using inertial measurements for propagation and visual ones for update, which adds priors from both measurements to the pose states.

\subsection{Local Mapping}
Every so often the current frame is selected to be a keyframe.
A new keyframe is made when the percent of features visible in the current frame compared to the most recent keyframe falls below a threshold.
Thus if the camera is always looking at the same thing, no new keyframe will be created.
Keyframes are pushed onto a local mapping window, but first a monocular depth network (MiDaS~\cite{Ranftl2022TPAMI, Ranftl2021ICCV}) is run on the raw image to obtain an affine-invariant inverse depth map which is up to a scale and bias parameter ($a$ and $b$, respectively).
All networks in AB-VINS are run on the raw distorted instead of undistorted image in order to benefit from the full available FoV.
The scale and bias, $a$ and $b$, as well as some other terms to correct the depth are estimated in the place of each feature position separately, which we call the AB feature representation.

\subsubsection{AB Features}
Given $a$ and $b$, and an affine-invariant inverse depth prediction $d_{inv}$ normalized into the [0, 1] range, we have $z_{inv} = a d_{inv} + b$, where $z_{inv}$ is the metric inverse depth.
To estimate $a$ and $b$ in a nonlinear optimization, we represent them as $s$ and $t$, where
\begin{align}
a &= a_{min} + (a_{max} - a_{min}) \mathrm{sigmoid}(s) \\
b &= b_{min} + (b_{max} - b_{min}) \mathrm{sigmoid}(t).
\label{eq:abst}
\end{align}
This makes it so that the scene scale can not become too large or too small in low-excitation scenarios, when scale is unobservable.
We use the DPT~\cite{Ranftl2021ICCV} Swin2~\cite{Liu2021ICCV} tiny transformer model open-sourced by MiDaS, which is applicable to embedded devices.

As shown in Fig.~\ref{fig:sys}, the watershed algorithm is then used on the edge image to create what we call a code mask, which partitions the image into $C$ different regions.
Using the code mask, for each keyframe a $C$-dimensional code vector $\mathbf{c}$ is estimated to push and pull different regions of the depth map in order to correct it using multi-view information.
Given a code mask element $m$ for $d_{inv}$ we can write 
\begin{align}
z_{inv} &= a (d_{inv} + c[m]) + b
\end{align}
where $x[i]$ indexes a vector $\mathbf{x}$.
$C$ in our system is 4.

After $\mathbf{c}$ of course comes $\mathbf{d}$.
We estimate a 4-dimensional vector $\mathbf{d}$ in order to make up for the fact that we use normalized bearings in AB-VINS, and run the MiDaS network on raw distorted images, while it was trained on a simple pinhole model (because the network predicts inverse depths that are too large near the edge of distorted images).
For $\mathbf{d}$ we write
\begin{align}
    z_{inv} &= a d_c (d_{inv} + c[m]) + b     \label{eq:d}\\
      d_c &= \theta_d / r  \\
    \theta_d &= \theta (1 + d[0] \theta^2 + d[1] \theta^4 + d[2] \theta^6 + d[3] \theta^8) \\
        r &= ||\mathbf{u}_n|| \\
        \theta &= \mathrm{atan2}(r, 1) 
\end{align}
where $\mathbf{u}_n$ is the ideal image coordinate, which is  a function $\mathbf{u}_n = \Pi^{-1}(\bm{\zeta}, \mathbf{u})$ of the camera intrinsics $\bm{\zeta}$ and raw image coordinate $\mathbf{u}$.
Thus $\mathbf{d}$ makes the inverse depth smaller closer to the edges of the image.
Since Eq.~\ref{eq:d} is nearly identical to the Kannala-Brandt~\cite{Kannala2006TPAMI} fisheye lens distorting equations, we call the process of $\mathbf{d}$ {\em fisheye monocular depth}.
The estimation of $a$ and $b$, with the optional $\mathbf{c}$ and $\mathbf{d}$ terms, to represent feature positions is called the AB feature representation.

After obtaining $z_{inv}$ we can write
\begin{align}
    {}^C\mathbf{p}_f &= \frac{\begin{bmatrix} \mathbf{u}_n^\top & 1 \end{bmatrix}^\top}{z_{inv}  \left|\left| {\begin{bmatrix} \mathbf{u}_n^\top & 1 \end{bmatrix}^\top} \right|\right|}
\label{eq:pf}
\end{align}
where ${}^C\mathbf{p}_f$ is the feature position in the camera frame.
The bearing $\begin{bmatrix} \mathbf{u}_n^\top & 1 \end{bmatrix}^\top$ has to be normalized to deal with largely distorted camera models.  
Using this formulation, we can estimate positions for {\em all} features on the image plane with only a few parameters.

\subsubsection{Depth Map Registration}
Before being pushed onto the local mapping window, the depth map is registered to the current local map -- linearly estimating $a$ and $b$ using the linear system presented in~\cite{Ranftl2022TPAMI} with the sparse points projected from the local map 
as the reference metric inverse depths.
Note that we wrap the linear registration in a RANSAC loop to improve the robustness.
Linear registration is followed by a nonlinear optimization to jointly estimate $a$, $b$, $\mathbf{c}$, and $\mathbf{d}$ while holding the local map fixed -- using the same 3D points as in tracking to static map for both linear and nonlinear registration.
The main cost in this optimization is the depth consistency cost, which is similar to the depth consistency costs in~\cite{Bloesch2018CVPR, Zuo2021ICRA} but defined in the inverse depth space in order for the cost to focus on closer objects rather than farther ones.
The $\mathbf{c}$ and $\mathbf{d}$ terms are initialized to zero, and a zero prior is placed on $\mathbf{c}$ and $\mathbf{d}$ in order to prevent the terms from blowing up, which is similar to the zero-code prior in~\cite{Bloesch2018CVPR, Zuo2021ICRA}.
In particular, the nonlinear depth map registration attempts to minimize the cost function $\mathcal{C}_{dr}$ defined as
\begin{align}
    \mathcal{C}_{dr} = \sum_{\ell \in \mathcal{V}_k} \lambda_\ell \rho_c(\frac{1}{\sigma_d^2}(r^{ds}_{\ell k})^2) + \norm{\mathbf{r}^p_k}^2_{\bm{\Sigma}_p}
\end{align}
where
\begin{align}
    r^{ds}_{\ell k} = \hat{z}_{inv_{\ell k}} - h_{ds}({}_A^{I_k}{\bar{q}}, {}^A{\mathbf{p}}_{I_k}, {}^A\mathbf{p}_{f_\ell}, {}^C_I\bar{q}, {}^C\mathbf{p}_{I}).
\end{align}
$\hat{z}_{inv_{\ell k}}$ is the estimated metric inverse depth of the $\ell$th feature at time $t_k$, and $h_{ds}$ projects the 3D point ${}^A\mathbf{p}_\ell$ into frame $k$ and extracts the inverse depth.
$\sigma_d$ is the manually-tuned noise  for the depth consistency residual.
We also have the zero $\mathbf{c}$, $\mathbf{d}$ prior:
\begin{align}
    \mathbf{r}^p_k = \mathbf{0} - \begin{bmatrix} \mathbf{c}_k^\top & \mathbf{d}_k^\top \end{bmatrix}^\top
\end{align}
which prevents $\mathbf{c}_k$ and $\mathbf{d}_k$ from becoming too large.
$\bm{\Sigma}_p$ is the manually-tuned diagonal covariance matrix for the zero prior.

\begin{figure}
\centering
\includegraphics[width=0.999\columnwidth,trim={28cm 2cm 3cm 6cm},clip]{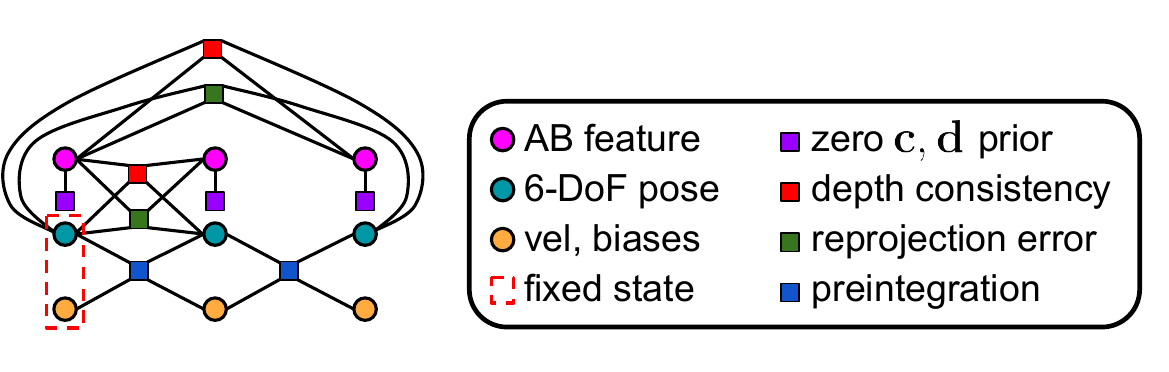}
\caption{A graphical representation of the local mapping optimization with three keyframes. For concise presentation, calibration states are not shown.}
\label{fig:local_ba}
\end{figure}

\subsubsection{Local Mapping Optimization} 
\label{sec:method:local_mapping:optimization}

After registration, as shown in Fig.~\ref{fig:local_ba} we perform a sliding-window optimization in the local mapper, 
with the oldest pose, velocity, and biases fixed. The default sliding window size of our system is 9.
Online calibration of the camera intrinsics and extrinsics is also performed in this optimization.

All of the poses are defined in the anchored frame $\{A\}$,
which is the gravity-aligned frame centered at the oldest keyframe and is changed after the oldest keyframe is popped from the window.
This is similar to robocentric VIO (R-VIO~\cite{Huai2019IJRR}), where the robocentric frame is defined to be the oldest frame  in the sliding window, which requires actually estimating the gravity direction.
On the other hand, our {\em anchored} frame is gravity-aligned -- only yaw-rotated when the oldest keyframe is popped off the window -- and does not required estimating gravity.
While the global pose is not tracked in the local mapper, AB-VINS still calculates it for system output by using the most recent global pose in the memory tree to transform poses in the $\{A\}$ frame into the global.\footnote{Note that traversal of the memory tree to calculate the global pose, which is logarithmic in the number of keyframes, is done solely in the mapping thread to keep VIO constant-time.}

Besides the standard reprojection error, our local mapping optimization also includes depth consistency cost.
Due to having a dense depth map for each keyframe, every bearing in the sliding window is actually its own landmark, and the landmarks simply have to work together with the oldest landmark via the depth consistency cost to keep the system stable.
Once the oldest bearing for a particular feature track is older than the sliding window, it becomes a static 3D landmark ${}^A\mathbf{p}_{f_\ell}$, and replaces the AB feature for the reprojection error and depth consistency.
More specifically, the local mapping optimization tries to minimize the follwoing cost function $\mathcal{C}_{lm}$:
\begin{align}
    \mathcal{C}_{lm} &= 
    \sum_{i \in \mathcal{K}} \sum_{\ell \in \mathcal{V}^s_i} \lambda_\ell \rho_c(\frac{1}{\sigma_d^2}(r^{ds}_{\ell i})^2)
    + \sum_{i \in \mathcal{K}} \sum_{\ell \in \mathcal{V}^s_i} \lambda_\ell 
    \rho_c\left(\norm{\mathbf{r}^{rs}_{\ell i}}^2_{\boldsymbol{\Sigma}_{v}}\right) \notag \\
    +& \sum_{i \in \mathcal{K}} \sum_{\ell \in \mathcal{V}^v_i} \lambda_\ell \rho_c(\frac{1}{\sigma_d^2}(r^{d}_{\ell i})^2)
    + \sum_{i \in \mathcal{K}} \sum_{\ell \in \mathcal{V}^v_i} \lambda_\ell 
    \rho_c\left(\norm{\mathbf{r}^{r}_{\ell i}}^2_{\boldsymbol{\Sigma}_{v}}\right) \notag \\
    +& \sum_{i \in \mathcal{K}} \norm{\hat{\mathbf{x}}_{i+1} \boxminus h_p(\hat{\mathbf{x}}_i, {}^i\boldsymbol{\alpha}_{i+1}, {}^i\boldsymbol{\beta}_{i+1})}^2_{\mathbf{Q}_{i,i+1}}
    + \sum_{i \in \mathcal{K}} \norm{\mathbf{r}^p_i}^2_{\bm{\Sigma}_p}
\end{align}
where $\mathcal{K}$ is the index set of all keyframes in the local mapping window, $\mathcal{V}^s_i$ is the index set of all 3D points that are older than the window visible in keyframe $i$, and $\mathcal{V}^v_i$ is the index set of all 3D points that are not older than the window visible in keyframe $i$.
Note that $\hat{\mathbf{x}}_i$ denotes an inertial state that is completely variable.
The residual $r^{d}_{\ell i}$, which is an extension of $r^{ds}_{\ell i}$ to include two AB features, is defined as
\begin{align}
    r^{d}_{\ell i} = \hat{z}_{inv_{\ell i}} - h_{d}({}_A^{I_k}\hat{\bar{q}}, {}^A\hat{\mathbf{p}}_{I_k}, \hat{z}_{inv_{\ell a}}, {}^C_I\hat{\bar{q}}, {}^C\hat{\mathbf{p}}_{I}, \hat{\bm{\zeta}}, \mathbf{u}_{\ell a})
\end{align}
where $\hat{z}_{inv_{\ell a}}$ is the metric inverse depth in the anchor frame for this 3D point at keyframe index $a$.
Similarly, $\mathbf{r}^{r}_{\ell i}$ is:
\begin{align}
    \mathbf{r}^{r}_{\ell i} = \mathbf{u}_{\ell i} - h_{r}({}_A^{I_k}\hat{\bar{q}}, {}^A\hat{\mathbf{p}}_{I_k}, \hat{z}_{inv_{\ell a}}, {}^C_I\hat{\bar{q}}, {}^C\hat{\mathbf{p}}_{I}, \hat{\bm{\zeta}}, \mathbf{u}_{\ell a})
\end{align}
which projects the 3D point from the inverse depth (parameterized by an AB feature) instead of using an existing static 3D point.
The same equation would be valid if we were estimating the inverse depth directly instead of the AB feature, as is standard.
Note that the camera extrinsics ${}^C_I\hat{\bar{q}}$, ${}^C\hat{\mathbf{p}}_{I}$ and intrinsics $\hat{\bm{\zeta}}$ are estimated where applicable in the local mapping optimization, including in the residuals $r^{ds}_{\ell i}$ and $\mathbf{r}^{rs}_{\ell i}$.
Due to the tight coupling of visual and inertial measurements in this optimization, the local map produced is scaled and gravity-aligned, and due to the sliding-window nature of the optimization, the computation does not grow over time.

\section{Mapping with the Memory Tree}
\label{method2}

After the oldest keyframe is popped off of the local mapping window, it is sent to the mapper thread, which is responsible for finding and solving for loop closures.
As shown in Fig.~\ref{fig:sys}, right before entering the mapper, the keyframe receives a global descriptor from NetVLAD~\cite{Arandjelovic2016CVPR} and local descriptors from SuperPoint~\cite{Detone18ArXiV}, which are typically different from the FAST corners used by VIO.
At this point, in order to save RAM memory, all image-resolution dense matrices (e.g., depth map and code mask) are deleted and only sparse portions are maintained at the local descriptor locations.
The global descriptor is PCA compressed to 512 dimensions, and searching for loop candidates with this descriptor is extremely efficient.
The local descriptors are used to match local keypoints between loop candidates, and if enough matches are found, Horn's method~\cite{Horn1988JOSA} wrapped in RANSAC is used to compute the relative pose for the pose graph optimization.
Upon a loop closure, the loop measurements are sent to the memory tree.

\begin{figure}
\centering
\includegraphics[width=0.999\columnwidth,trim={5cm 1.25cm 33.5cm 5.5cm},clip]{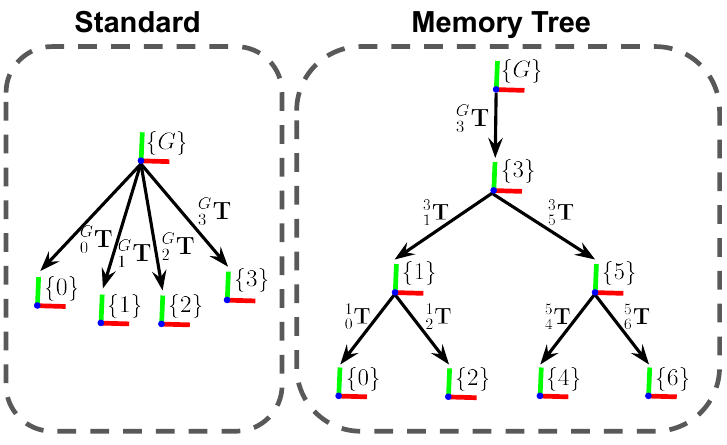}
\caption{A comparison of the pose definitions in standard factor graphs (left) and our memory tree (right). In the memory tree, poses are defined in the parent frame for nodes within a binary search tree rather than all in one global frame.
}
\label{fig:mem_tree_def}
\end{figure}

\subsection{Tree Structure}

The memory tree is a {\em novel} data structure proposed to significantly speed up pose graph optimization.
As seen in Fig.~\ref{fig:mem_tree_def}, poses in the memory tree are defined in the parent frame within a binary search tree,
while only the root's pose is in the global frame.
As VINS has 4 unobservable directions~\cite{Yang2023TRO} and drifts in 3D position and yaw, we adopt  the 4-DoF poses  in the memory tree:
\begin{align}
{}_j^i\mathbf{T} &= \begin{bmatrix} \mathbf{R}_z(-{}_i^j\theta_z) & {}^i\mathbf{p}_j\\ \mathbf{0}_{1 \times 3} & 1 \end{bmatrix}
\label{eq:4dof}
\end{align}
where ${}_i^j\theta_z$ is the yaw angle, $\mathbf{R}_z$ constructs a yaw-only 3D rotation matrix, and ${}^i\mathbf{p}_j$ is the gravity-aligned position.
The keyframe's timestamp is the search key for the tree.
The memory tree is balanced using the AVL tree rotation rules~\cite{Adelsonvelskii1963TR}, which guarantee a perfectly balanced tree no matter what.
Since the tree is always balanced, calculating the global pose for any keyframe can be done in $\mathcal{O}(\log(N))$ time if there are $N$ nodes in the tree by repeatedly transforming the pose into the same frame as the parent's pose until the root node is reached.
The benefit of defining the poses in this way is that optimizing a single pose also moves the entire subtree underneath the optimized pose's node.

To limit the memory usage in areas of repeated visitation, redundant nodes are pruned from the memory tree.
A node is considered redundant if a certain percent of the map points visible in its keyframe are visible in at least three other keyframes, which is similar to the keyframe culling criteria of ORB-SLAM~\cite{Campos2021TRO}.
After a node is pruned, the tree is rebalanced, again using the AVL balancing rules, to ensure that the height of the tree is $\mathcal{O}(\log(N))$.

\begin{figure*}
\centering
\includegraphics[width=0.888\textwidth,trim={0cm 0cm 9.25cm 5.5cm},clip]{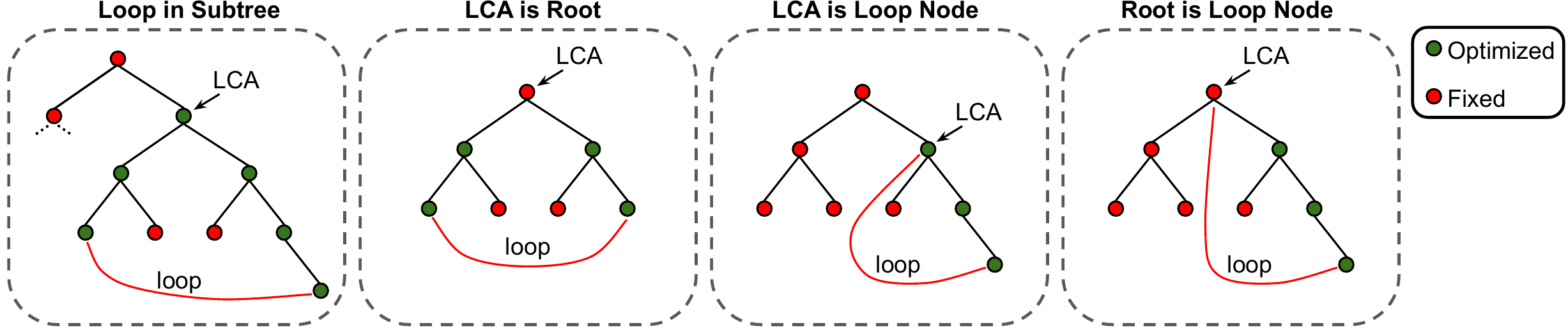}
\caption{Examples of different optimization patterns for the memory tree when optimizing the full path between loop nodes.}
\label{fig:mem_tree_examples}
\end{figure*}

\begin{figure}
\centering
\includegraphics[width=0.999\columnwidth,trim={0cm 0cm 24.5cm 5.5cm},clip]{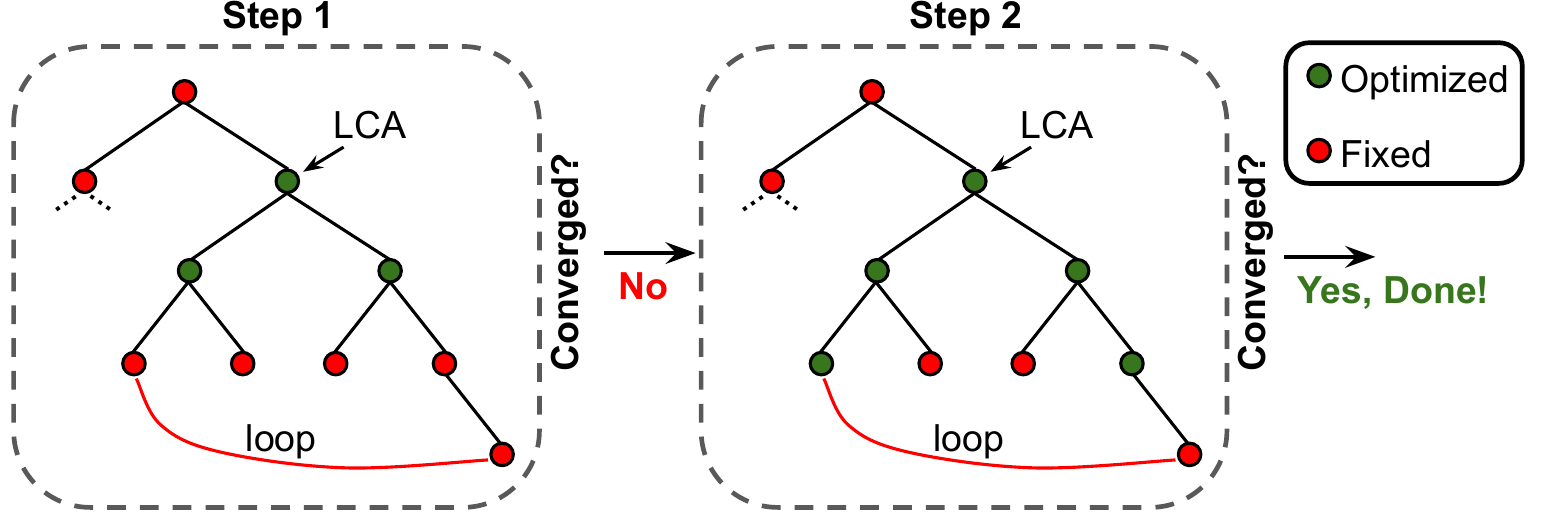}
\caption{The process of top-down memory tree optimization. The LCA and its children along the path between the two loop nodes are set variable from the top down until convergence.}
\label{fig:mem_tree_top_down}
\end{figure}

\subsection{Optimizing Memory Tree}

Optimizing the memory tree is straightforward.
We wish to minimize the cost function $\mathcal{C}_{tree}$ defined as
\begin{align}
\mathcal{C}_{tree} = \sum_{i \in \mathcal{S}} \rho_c\left(\norm{\mathbf{r}^{tree}_{i,i+1}}^2_{\boldsymbol{\Sigma}_{tree}}\right)
+ \sum_{i,j \in \mathcal{L}} \rho_c\left(\norm{\mathbf{r}^{tree}_{ij}}^2_{\boldsymbol{\Sigma}_{tree}}\right)
\label{eq:tree_cost}
\end{align}
where $\mathcal{S}$ is the set of necessary sequential connections to add (nodes that are next to each other temporally) and $\mathcal{L}$ is the set of necessary loop connections to add.
$\boldsymbol{\Sigma}_{tree}$ is the manually-tuned diagonal covariance matrix for tree edges, and $\rho_c$ is again the Cauchy loss function.
Given a 4-DoF relative pose measurement ${}^{i}_{j}\theta_z$, ${}^{i}\mathbf{p}_{j}$ between nodes $i$ and $j$, the residual is
\begin{align}
\mathbf{r}^{tree}_{ij} = \begin{bmatrix}
{}^{i}_{j}\theta_z  - ({}^{i}_{L}\hat{\theta}_z - {}^{j}_{L}\hat{\theta}_z)
\\ 
{}^{i}\mathbf{p}_{j} - \mathbf{R}_z({}^{i}_{L}\hat{\theta}_z)({}^{L}\hat{\mathbf{p}}_{j} - {}^{L}\hat{\mathbf{p}}_{i})
\end{bmatrix} \label{eq:tree_edge}
\end{align}
where $\{i\}$, $\{j\}$ are the local gravity-aligned frames for nodes $i$ and $j$, and $\{L\}$ is the gravity-aligned frame centered around the lowest common ancestor (LCA) for nodes $i$ and $j$.
Finding the LCA can be performed in $\mathcal{O}(\log(N))$ time.
Of course the poses for the nodes are defined in the parent frame and not the $\{L\}$ frame, but the poses in the $\{L\}$ frame have to be found by transforming iteratively into the same frame as the parent until the LCA is reached.
If optimizing all states in the memory tree, the Hessian structure will be very dense due to correlating more states than usual.
Of course, the point of the memory tree is not to optimize every state, but to carefully choose which states to optimize.

Given two loop candidate tree nodes, we can actually move all the nodes needed to close the loop by only optimizing nodes along the shortest path between the two loop nodes.
In order to do this, first the LCA must be found between the two.
This is because the shortest path between any two nodes in the tree always passes through the LCA.
To stabilize the optimization, the root node always remains fixed.
Note that this does in most cases allow the first ever pose to move, but we argue that the global pose can be simply taken relative to the first pose.
Fig.~\ref{fig:mem_tree_examples} shows different examples of optimization patterns on the memory tree when optimizing the full path between two loop nodes.
Note that edges with only fixed parameters need not be included in the optimization.
Thus, not all measurements are added to the optimization problem.
The path between the two loop nodes is traversed, and only at most $\mathcal{O}(\log(N))$ sequential relative edges are added, and all loop measurements are checked to see if any of their parameters are included in the set of variable parameters.
If so, the loop measurement is added, and if not it is left out of the optimization problem.

While optimizing the full path between two loop nodes is already highly-efficient, an even more efficient solution exists.
Since optimizing a single node's pose moves the entire subtree underneath, it is possible to start with only optimizing the LCA and its two children, and work our way down until the optimization converges in one iteration.
We call this the memory tree top-down optimization algorithm.
Fig.~\ref{fig:mem_tree_top_down} shows how this works.
This solution is not only highly-efficient, but it also solves pose graph SLAM while only relinearizing a {\em constant} number of variables. 
This is because on average only three states are variable (the LCA and its two children), while the rest are fixed.
However, in the worst case $\mathcal{O}(\log(N))$ states are set variable.

To robustify against potential bad loop closure measurements (e.g., from visual aliasing), we propose a robust yet efficient $\chi^2$ check.
Before optimizing the memory tree with a new loop measurement, first the 4-DoF states of all the nodes along the shortest path between the two loop nodes are copied and stored.
After optimizing the memory tree, if the new loop residual does not pass a $\chi^2$ test with theoretical threshold $\gamma$ (i.e., $||\mathbf{r}^{tree}_{ij}||^2_{\boldsymbol{\Sigma}_{tree}} \geq \gamma$ if the newest loop measurement is between nodes $i$ and $j$), then the old states are copied back into the memory tree to restore it to what it was before, and the loop measurement is discarded.
As such, the proposed AB-VINS is robust to catastrophically bad loop closures while only using $\mathcal{O}(\log(N))$ extra memory.

\subsection{Complexity Analysis}

\begin{table*} \centering
\caption{
Algorithm analysis of different pose graph optimization methods with $N$ keyframes.
}
\label{tab:pose_graph_analysis}
\resizebox{0.99\textwidth}{!}{\begin{tabular}{@{}c|ccc|ccc|ccc@{}}
\toprule
 & \multicolumn{3}{|c}{\textbf{Num. Variables Affected}} & \multicolumn{3}{|c}{\textbf{Calculate Global Pose}} & \multicolumn{3}{|c}{\textbf{Optimization}} \\ 
\midrule
Method & Best Case & Worst Case & Avg. Case & Best Case & Worst Case & Avg. Case & Best Case & $\mathcal{O}(1)$ Loops & $\mathcal{O}(N)$ Loops \\
\midrule
Standard      & $\Theta(N)$ & $\Theta(N)$ & $\Theta(N)$ & $\boldsymbol{\Theta(1)}$  & $\boldsymbol{\Theta(1)}$  & $\boldsymbol{\Theta(1)}$ & $\Omega(N)$ & $\mathcal{O}(N)$ & $\mathcal{O}(N^3)$ \\
iSAM2~\cite{Kaess2011IJRR}      & $\boldsymbol{\Omega(1)}$ & $\mathcal{O}(N)$ & $\mathcal{O}(N)^\dag$ & $\boldsymbol{\Theta(1)}$  & $\boldsymbol{\Theta(1)}$  & $\boldsymbol{\Theta(1)}$ & $\boldsymbol{\Omega(1)}$ & $\mathcal{O}(N)$ & $\mathcal{O}(N^3)$ \\
Relative~\cite{Sibley2010RSS}   & $\boldsymbol{\Omega(1)}$ & $\mathcal{O}(N)$ & $\boldsymbol{\mathcal{O}(1)}$ & $\boldsymbol{\Omega(1)}$ & $\mathcal{O}(N)$ & $\mathcal{O}(N)$ & $\boldsymbol{\Omega(1)}^*$ & $\mathcal{O}(N^3)$ & $\mathcal{O}(N^3)$ \\
Mem. Tree All States          & $\Theta(N)$ & $\Theta(N)$ & $\Theta(N)$ & $\boldsymbol{\Omega(1)}$ & $\mathcal{O}(\log(N))$ & $\mathcal{O}(\log(N))$ & $\Theta(N^3)$ & $\Theta(N^3)$ & $\Theta(N^3)$ \\
Mem. Tree Full Path           & $\boldsymbol{\Omega(1)}$ & $\boldsymbol{\mathcal{O}(\log(N))}$ & $\mathcal{O}(\log(N))^\dag$ & $\boldsymbol{\Omega(1)}$ & $\mathcal{O}(\log(N))$ & $\mathcal{O}(\log(N))$ & $\boldsymbol{\Omega(1)}$ & $\boldsymbol{\mathcal{O}((\log(N))^3)}$ & $\boldsymbol{\mathcal{O}(N\log(N))}$ \\
Mem. Tree Top-Down           & $\boldsymbol{\Omega(1)}$ & $\boldsymbol{\mathcal{O}(\log(N))}$ & $\boldsymbol{\mathcal{O}(1)}$ & $\boldsymbol{\Omega(1)}$ & $\mathcal{O}(\log(N))$ & $\mathcal{O}(\log(N))$ & $\boldsymbol{\Omega(1)}$ & $\boldsymbol{\mathcal{O}((\log(N))^3)}$ & $\boldsymbol{\mathcal{O}(N\log(N))}$ \\
\bottomrule
\multicolumn{10}{l}{${}^\dag$\footnotesize{Assuming a loop-closure scenario. Number of variables affected is constant under exploration.}} \\
\multicolumn{10}{l}{\textsuperscript{*}\footnotesize{Complexity would be $\Omega(N)$ if following the method in the original paper~\cite{Sibley2010RSS} (i.e., including every measurement instead of just the necessary ones).}} \\
\end{tabular}}
\end{table*}

Table~\ref{tab:pose_graph_analysis} shows the complexity analysis of different pose graph optimization algorithms when optimizing with $N$ keyframes.
It is evident that our memory tree top-down method is the overall best.
Looking at the number of variables affected by optimization (i.e., the number of variables that need to be relinearized), the standard method as in SPA~\cite{Konolige2010IROS} and iSAM~\cite{Kaess2008TRO}, with the poses defined in the global frame, affects $\Theta(N)$ variables no matter what.
iSAM2~\cite{Kaess2011IJRR} has a chance to affect a constant number of variables under exploration, but typically affects $\mathcal{O}(N)$ variables under a loop closure scenario.
The relative method~\cite{Sibley2010RSS} can affect up to $\mathcal{O}(N)$ variables, but on average can affect only a constant number of variables due to only optimizing the beginning and end of each loop.
The memory tree top-down method can also affect a constant number of variables, but in the worst case is only $\mathcal{O}(\log(N))$, and only affects a constant number of variables on average even under loop closure scenarios.
Moving on to the global pose calculation, defining the poses in the global frame as is standard allows for constant-time global pose calculation.
For the relative method, global pose calculation typically takes $\mathcal{O}(N)$ time since each keyframe pose is defined in the previous one's frame of reference, and thus the global pose is dependent on each previous keyframe.
On the other hand, with the memory tree, global pose calculation is typically done in $\mathcal{O}(\log(N))$ time due to the efficient tree structure of the pose definitions.
It should be noted that the computation incurred by having to relinearize a larger number of variables is much higher than having to iterate to find the global pose.

Table~\ref{tab:pose_graph_analysis} also shows the computational complexity of the overall optimization.
Note that an average case is not reported here since the complexity depends on the number of loop measurements, which depends on the environment.
Thus, there are two worst cases reported for $\mathcal{O}(1)$ and $\mathcal{O}(N)$ loop measurements.
With a standard pose graph optimization method, the complexity is linear in the absolute best case, which is the same as having a constant number of loops.
In the worst case with $\mathcal{O}(N)$ loops, the complexity is $\mathcal{O}(N^3)$~\cite{Konolige2010IROS}.
For iSAM2~\cite{Kaess2011IJRR}, the optimization can be constant-time in the best case (under exploration) but other than that shares the same complexity with the standard pose graph optimization.
With the relative method, the best case arises when the two loop keyframes are very close to each other in time, and there are a constant number of loops.
In this case, the method can be constant time.
In the worst case with a constant number of loops, the complexity is cubic.
This is because $\mathcal{O}(N)$ states can be set variable even with a single loop, and the states are all correlated leading to a dense Hessian structure.
Similarly, the complexity of the relative method is cubic with $\mathcal{O}(N)$ loops.
Note that this analysis of the relative method assumes that 1) it is applied to pose graph optimization and 2) only the necessary measurements are included in the optimization instead of all the measurements as proposed in the original paper~\cite{Sibley2010RSS}.
If all measurements are included in the optimization then the best case becomes $\Omega(N)$ due to having to iterate through all of the sequential edges.

Moving on the the memory tree, with all states set variable, the complexity is cubic in every case.
This is due to the cost of LM iterations with a nearly dense Hessian structure.
For the full path algorithm, the best case is when there are a constant number of loops and the two loop nodes are near the root of the tree.
In this case, the complexity is constant.
In the worst case with a constant number of loops, the complexity is $\mathcal{O}((\log(N))^3)$, which is sublinear.
This is due to the cost of LM iterations with $\mathcal{O}(\log(N))$ states that are all correlated.
In the worst case with $\mathcal{O}(N)$ loops, the complexity is $\mathcal{O}(N\log(N))$ due to having to find the LCA for $\mathcal{O}(N)$ loops to see which measurements should be included in the optimization (which is typically more efficient than the LM iterations).
For the top-down method, the complexity is the same as the full path algorithm since $\mathcal{O}(\log(N))$ states can be variable in the worst case.
If only a constant number of states are variable (which is the average case) with a constant number of loops,
the complexity is $\mathcal{O}((\log(N))^2)$ due to having to find the LCA for the $\mathcal{O}(\log(N))$ relative edges.
Since AB-VINS prunes many duplicate nodes, loop measurements are typically discarded, leading to a constant number of loop measurements.
While AB-VINS allows for at most $\mathcal{O}(N)$ loop measurements, some systems such as ORB-SLAM3~\cite{Campos2021TRO} allow for up to $\mathcal{O}(N^2)$ loops.
If this is the case, the worst case for the full path and top-down algorithms becomes $\mathcal{O}(N^2\log(N))$, which is still better than the cubic complexity of other methods.

\section{Experimental Results} 
\label{sec:experiments}

We implement the proposed AB-VINS in an efficient C++ codebase.
Ceres solver~\cite{Agarwal2022Ceres} with single-thread execution is used for all optimizations.
Automatic differentiation provided by \texttt{ceres::Jet} is used in all factors except for preintegration and the reprojection error from a static point used in tracking to static map, which use analytical Jacobians.
Unlike numerical differentiation via finite differences, automatic differentiation does not come with a large computational overhead or loss of accuracy.
We found automatic differentiation to be especially useful for developing the memory tree relative pose factor, since the analytical Jacobians for it are complicated.
A desktop  with an i5-6600K CPU and RTX 2070 Super GPU with 16GB of RAM is used in all experiments.

\begin{table*} \centering
\caption{
ATE (deg/m) on the AR Table dataset.
}
\label{tab:ar_table_ate}
\resizebox{0.99\textwidth}{!}{\begin{tabular}{@{}cccccccccccc@{}}
\toprule
\textbf{Type} & \textbf{Algorithm}
 &\textbf{table 1} & \textbf{table 2} & \textbf{table 3} & \textbf{table 4} & \textbf{table 5} & \textbf{table 6} & \textbf{table 7} & \textbf{table 8} & \textbf{Average} \\ \midrule
 \multirow{6}{*}{VIO} 
& OKVIS~\cite{Leutenegger2014IJRR} & 1.769 / 0.096 & 1.472 / 0.061 & 3.519 / 0.123 & 0.919 / 0.139 & 0.667 / 0.055 & 1.062 / 0.078 & 3.373 / 0.166 & 2.096 / 0.197 & 1.860 / 0.114 \\
& OpenVINS~\cite{Geneva2020ICRA} & 1.019 / 0.059 & 0.984 / 0.032 & 1.309 / 0.042 & 0.650 / 0.043 & 0.914 / 0.032 & 1.218 / 0.043 & 1.055 / 0.054 & 0.965 / 0.084 & \underline{1.014} / \bf{0.049} \\
& AB-VINS VIO & 4.536 / 0.264 & 1.577 / 0.475 & 5.318 / 0.553 & 3.070 / 0.354 & 5.379 / 0.315 & 4.033 / 0.380 & 3.994 / 0.899 & 5.965 / 0.882 & 4.234 / 0.515 \\
& AB-VINS VIO no $\mathbf{c}$ & 9.061 / 0.682 & 2.053 / 2.434 & 10.286 / 0.803 & 7.749 / 0.956 & 18.582 / 1.705 & 8.545 / 1.525 & 4.837 / 1.008 & 9.833 / 2.719 & 8.868 / 1.479 \\
& AB-VINS VIO no $\mathbf{d}$ & 8.465 / 0.853 & 3.004 / 2.454 & 11.916 / 1.280 & 4.996 / 1.059 & 11.652 / 0.975 & 7.094 / 1.114 & 6.741 / 1.163 & 3.714 / 2.644 & 7.198 / 1.443 \\
& AB-VINS VIO + Depth Sens. & 4.179 / 0.129 & 1.428 / 0.039 & 2.011 / 0.165 & 3.572 / 0.182 & 1.488 / 0.088 & 2.410 / 0.098 & 2.699 / 0.344 & 5.017 / 0.592 & 2.850 / 0.205 \\
\midrule
 \multirow{3}{*}{SLAM} 
 & VINS-Fusion~\cite{Qin2018TRO} & 0.631 / 0.042 & 1.006 / 0.106 & 0.421 / 0.025 & 0.658 / 0.043 & 0.653 / 0.019 & 0.609 / 0.064 & 0.814 / 0.036 & 0.662 / 0.053 & \bf{0.682} / \bf{0.049} \\
 & ORB-SLAM3~\cite{Campos2021TRO} & 2.385 / 0.079 & 3.248 / 0.067 & 2.584 / 0.048 & 2.601 / 0.116 & 2.180 / 0.113 & 1.543 / 0.127 & 2.065 / 0.187 & 1.003 / 0.145 & 2.201 / \underline{0.110} \\
 & AB-VINS & 4.445 / 0.292 & 3.449 / 0.628 & 4.816 / 0.417 & 3.869 / 0.369 & 6.528 / 0.281 & 4.712 / 0.389 & 7.335 / 0.787 & 3.634 / 0.693 & 4.849 / 0.482 \\
\bottomrule
\end{tabular}}
\end{table*}

\begin{table*} \centering
\caption{
ATE (deg/m) on the TUM-VI dataset.
}
\label{tab:tum_vi_ate}
\begin{tabular}{@{}cccccccccc@{}}
\toprule
\textbf{Type}  & \textbf{Algorithm}
 & \textbf{room 1} & \textbf{room 2} & \textbf{room 3} & \textbf{room 4} & \textbf{room 5} & \textbf{room 6} & \textbf{Average} \\ \midrule
\multirow{3}{*}{VIO}
& OKVIS & 1.007 / 0.050 & 0.776 / 0.126 & 1.092 / 0.069 & 0.959 / 0.039 & 1.854 / 0.064 & 0.584 / 0.047 & 1.045 / 0.066 \\
& OpenVINS & 1.263 / 0.058 & 2.395 / 0.075 & 1.362 / 0.076 & 0.845 / 0.041 & 1.423 / 0.083 & 0.754 / 0.032 & 1.340 / 0.061 \\
& AB-VINS VIO & 1.987 / 0.508 & 5.737 / 3.260 & 6.110 / 1.412 & 1.347 / 0.190 & 3.160 / 0.627 & 1.583 / 0.174 & 3.321 / 1.028 \\
\midrule
\multirow{3}{*}{SLAM}
& VINS-Fusion & 0.760 / 0.046 & 1.419 / 0.104 & 0.698 / 0.051 & 0.567 / 0.026 & 1.701 / 0.063 & 0.698 / 0.048 & \underline{0.974} / \underline{0.057} \\
& ORB-SLAM3 & 0.482 / 0.012 & 0.607 / 0.015 & 1.315 / 0.038 & 0.481 / 0.014 & 0.531 / 0.014 & 0.465 / 0.013 & \bf{0.647} / \bf{0.018} \\
& AB-VINS & 4.615 / 0.311 & 5.092 / 0.462 & 4.331 / 0.181 & 1.714 / 0.112 & 4.372 / 0.466 & 1.294 / 0.190 & 3.570 / 0.287 \\
\bottomrule
\end{tabular}\end{table*}

\begin{figure*}
\centering
\includegraphics[width=0.24\textwidth,trim={2.7cm 1.2cm 4.2cm 0.7cm}, clip]{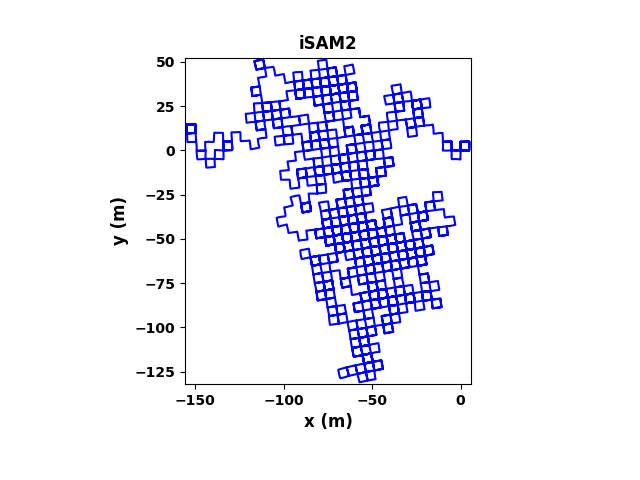}
\includegraphics[width=0.24\textwidth,trim={2.7cm 1.2cm 4.2cm 0.7cm}, clip]{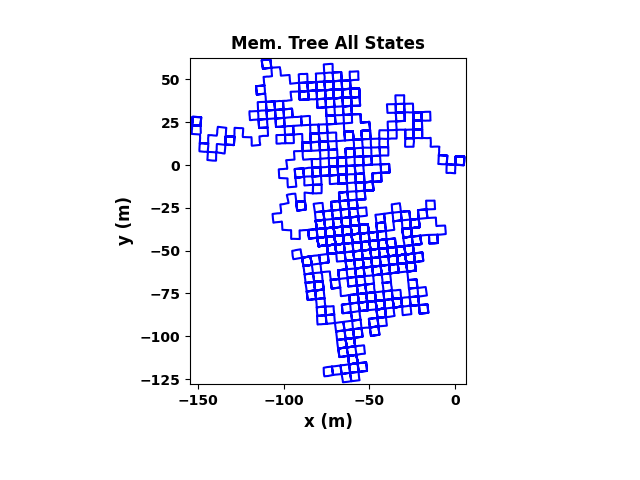}
\includegraphics[width=0.24\textwidth,trim={2.7cm 1.2cm 4.2cm 0.7cm}, clip]{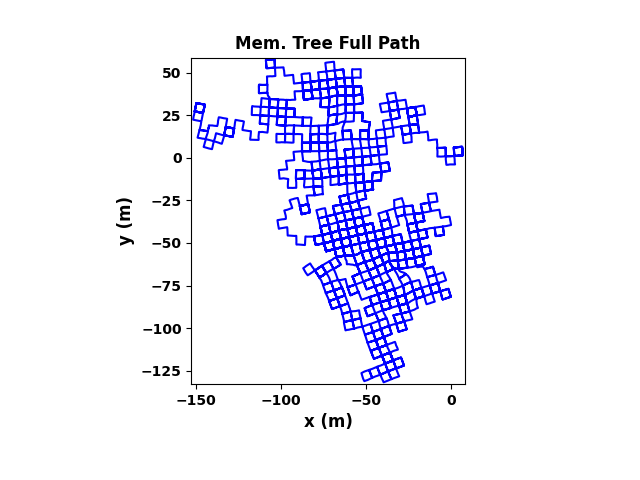}
\includegraphics[width=0.24\textwidth,trim={2.7cm 1.2cm 4.2cm 0.7cm}, clip]{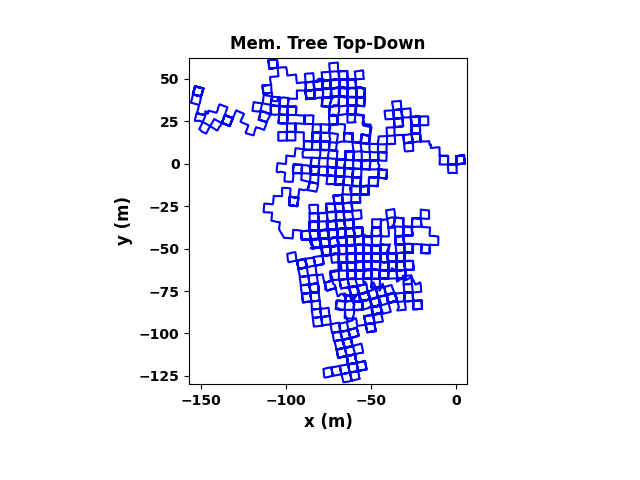}
\caption{The trajectory results of different incremental pose graph optimization methods on the M10000 dataset.
From left to right: iSAM2~\cite{Kaess2011IJRR}, memory tree with all states optimized, memory tree full path, and memory tree top-down.
}
\label{fig:tree_opt_traj_M10000}
\end{figure*}

\begin{figure*}
\centering
\includegraphics[width=0.24\textwidth,trim={.2cm 1.2cm 1.5cm .5cm}, clip]{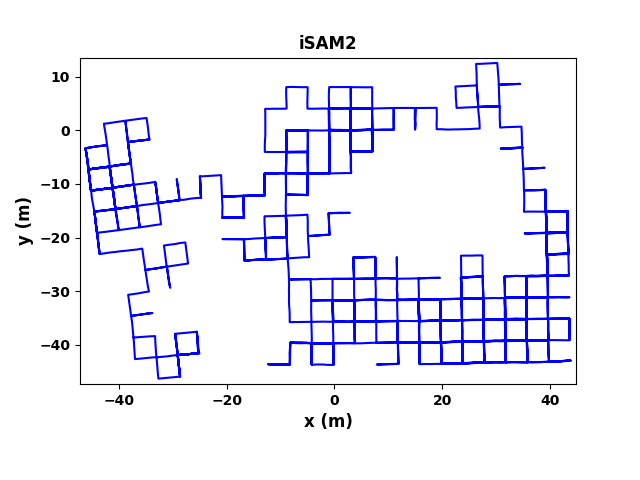}
\includegraphics[width=0.24\textwidth,trim={.2cm 1.2cm 1.5cm .5cm}, clip]{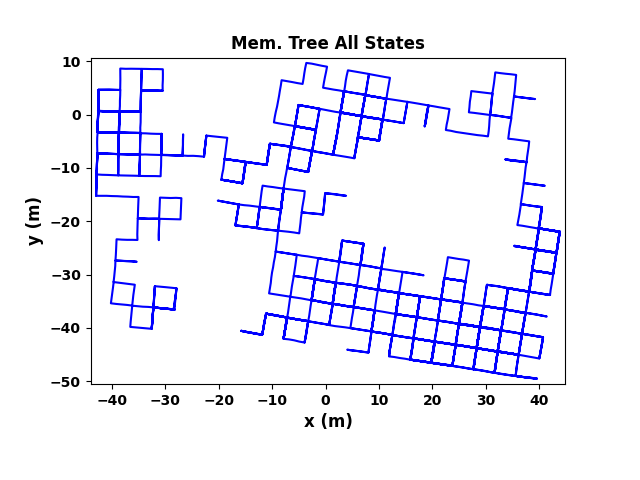}
\includegraphics[width=0.24\textwidth,trim={.2cm 1.2cm 1.5cm .5cm}, clip]{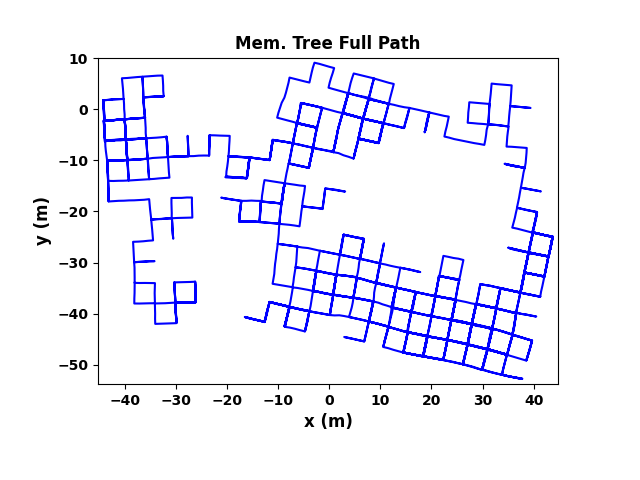}
\includegraphics[width=0.24\textwidth,trim={.2cm 1.2cm 1.5cm .5cm}, clip]{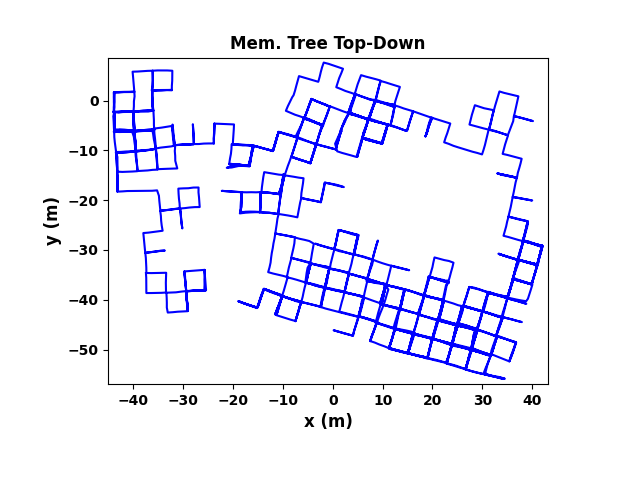}
\caption{The trajectory results of different incremental pose graph optimization methods on the M3500 dataset.
From left to right: iSAM2~\cite{Kaess2011IJRR}, memory tree with all states optimized, memory tree full path, and memory tree top-down.
}
\label{fig:tree_opt_traj_M3500}
\end{figure*}

\begin{figure*}
\centering
\includegraphics[width=0.24\textwidth,trim={2.7cm 1.2cm 3.5cm 0.7cm}, clip]{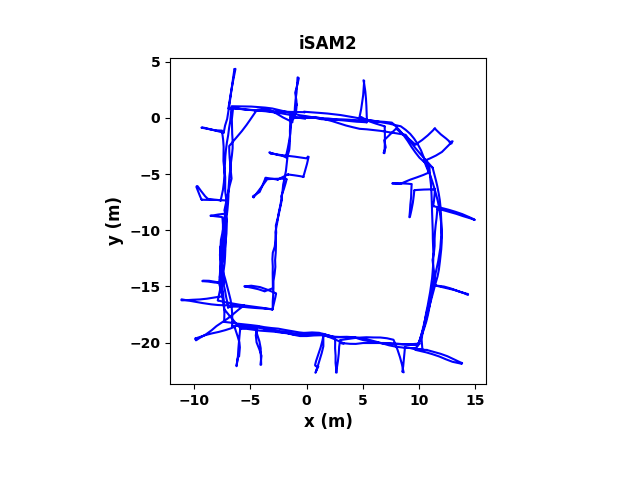}
\includegraphics[width=0.24\textwidth,trim={2.7cm 1.2cm 3.5cm 0.7cm}, clip]{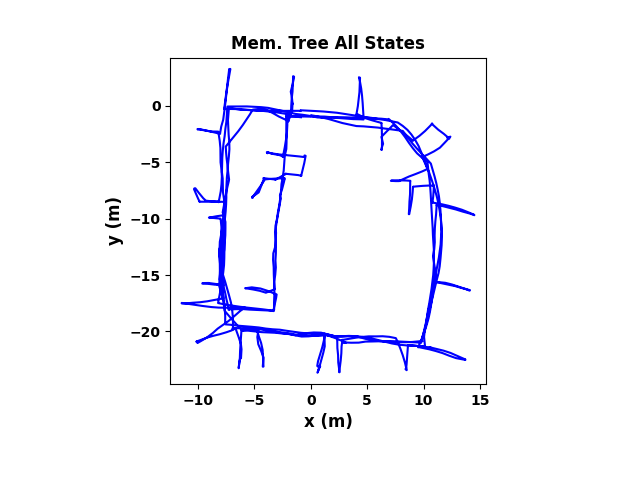}
\includegraphics[width=0.24\textwidth,trim={2.5cm 1.2cm 3.5cm 0.7cm}, clip]{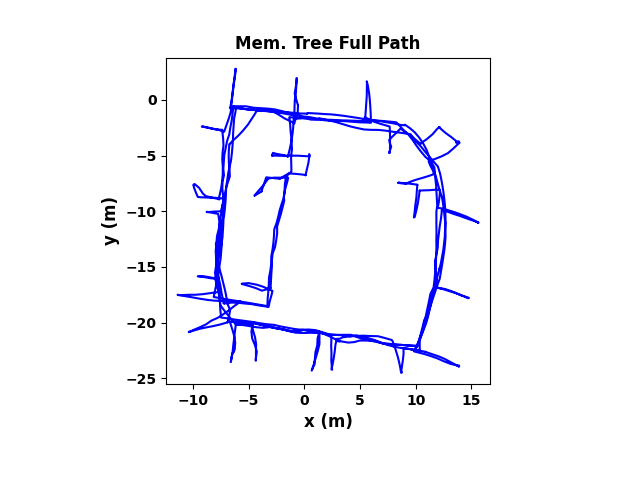}
\includegraphics[width=0.24\textwidth,trim={2.7cm 1.2cm 3.5cm 0.7cm}, clip]{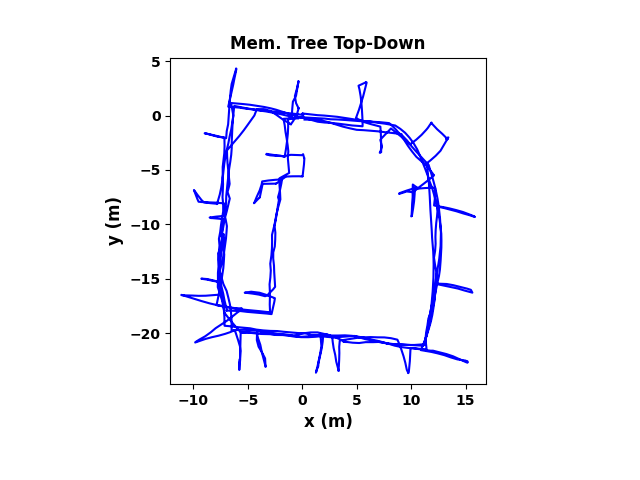}
\caption{The trajectory results of different incremental pose graph optimization methods on the Intel dataset.
From left to right: iSAM2~\cite{Kaess2011IJRR}, memory tree with all states optimized, memory tree full path, and memory tree top-down.
}
\label{fig:tree_opt_traj_INTEL}
\end{figure*}

\begin{figure*}
\centering
\includegraphics[height=4.7cm,trim={1.8cm 1.2cm 3cm 0.7cm}, clip]{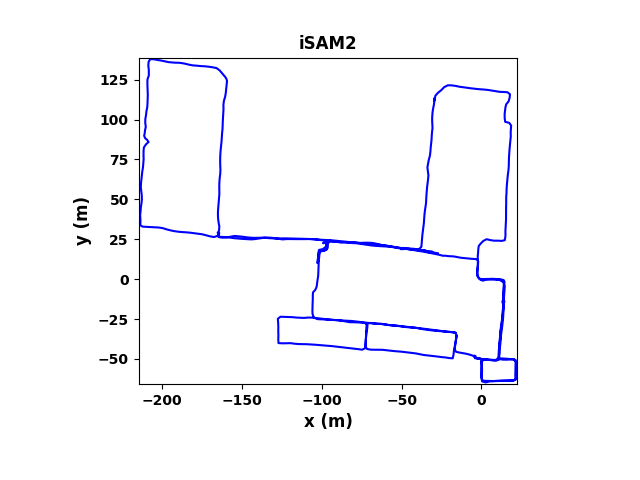}
\includegraphics[height=4.7cm,trim={3cm 1.3cm 4.8cm 0.7cm}, clip]{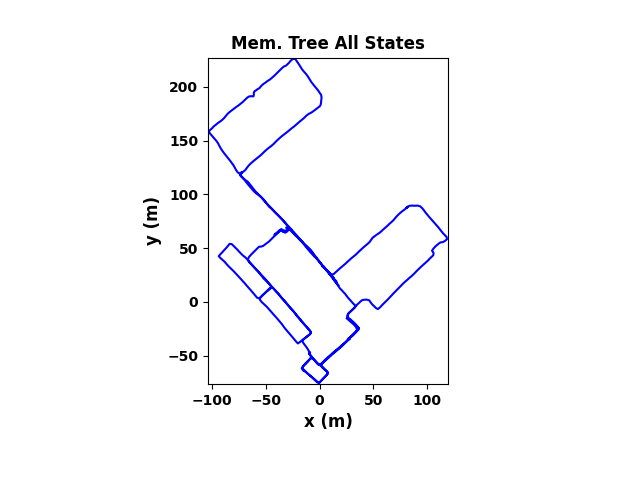}
\includegraphics[height=4.7cm,trim={3cm 1.3cm 4.5cm 0.7cm}, clip]{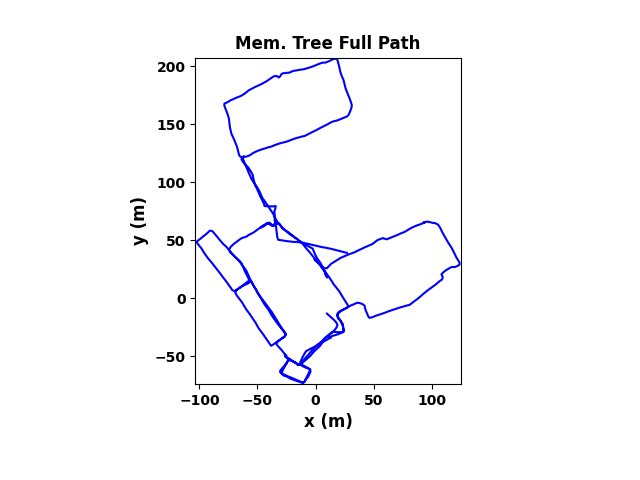}
\includegraphics[height=4.7cm,trim={3cm 1.3cm 4.2cm 0.7cm}, clip]{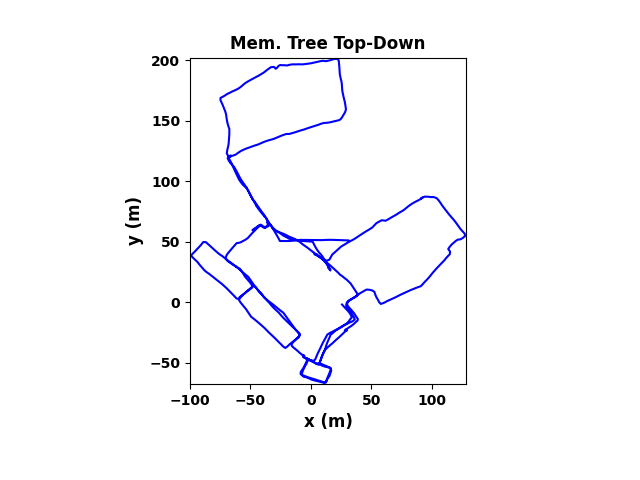}
\caption{The trajectory results of different incremental pose graph optimization methods on the MIT Killian Court dataset.
From left to right: iSAM2~\cite{Kaess2011IJRR}, memory tree with all states optimized, memory tree full path, and memory tree top-down.
}
\label{fig:tree_opt_traj_MIT}
\end{figure*}

\subsection{Accuracy} \label{sec:accuracy}

We first evaluate the localization accuracy of the proposed AB-VINS by comparing to the state-of-the-art VIO~\cite{Leutenegger2014IJRR,Geneva2020ICRA} and visual-inertial SLAM~\cite{Qin2018TRO,Campos2021TRO} methods.
Trajectories are aligned by finding the best 4-DoF transformation between the estimated and ground truth,
and the Absolute Trajectory Error (ATE) is used as the performance metric.

\subsubsection{Pose Accuracy on the AR Table Dataset} \label{sec:ar_table}

We first test on the recently released AR Table dataset~\cite{Chen2023ICRA}, which has a single monocular camera with radial and tangential distortion.
The results are shown in Table~\ref{tab:ar_table_ate}, showing that our AB-VINS is not as accurate as the state-of-the-art VINS.
An ablation study is presented in the table where we investigate the impact of fixing each keyframe's $\mathbf{c}$ and $\mathbf{d}$ in the local mapping window to zero separately.
Clearly it is necessary to estimate both $\mathbf{c}$ and $\mathbf{d}$ to correct the monocular depth.
Additionally, we  evaluate the impact of utilizing a depth sensor instead of monocular depth.
Note that $a$ and $b$ are fixed to 1 and 0 when using the depth sensor.
There is a clear improvement from using more accurate depth from the depth sensor rather than the monocular depth, 
which shows that the main reason for AB-VINS being less accurate than the other systems is the reliance on monocular depth.
Interestingly, the loop closures of AB-VINS (using the memory tree)  slightly improve position accuracy while degrading orientation  in this case, which at least shows that the memory tree did not hurt the overall accuracy.

\subsubsection{Pose Accuracy on the TUM-VI Dataset} \label{sec:tum_vi}

The next dataset tested on is the TUM-VI dataset~\cite{Schubert2018IROS}.
We use the left fisheye camera for evaluation,
and all the hyperparameters of AB-VINS are the same as in the AR Table dataset.
The results are reported in Table~\ref{tab:tum_vi_ate}.
Although the AB-VINS is again not as accurate as the state-of-the-art,
the accuracy of AB-VINS with loop closure is significantly higher than the VIO  on this dataset,
showing the capability of the memory tree to improve the accuracy over simply running VIO.

\subsubsection{Depth Accuracy}

\begin{table} \centering
\caption{
Depth accuracy on the AR Table dataset.
}
\label{tab:ar_table_depth}
\resizebox{0.99\columnwidth}{!}{\begin{tabular}{@{}cccccccc@{}}
\toprule
 & \textbf{Method} & $\bm\delta_{1} \uparrow$ & $\bm\delta_{2} \uparrow$ & $\bm\delta_{3}\uparrow$ & \textbf{RMSE} $\downarrow$  & \textbf{MAE} $\downarrow$ & \textbf{log}${}_{10} \downarrow$ \\
\midrule
\multirow{2}{*}{table 1}
& MiDaS & \bf 0.733 & 0.949 &  0.991 &  0.785 & \bf  0.124 & \bf  0.072 \\
& AB & 0.724 & \bf 0.954 & \bf 0.992 & \bf 0.756 & 0.129 & 0.075 \\
\midrule
\multirow{2}{*}{table 2}
& MiDaS & 0.184 & 0.375 & 0.627 & 1.231 & 0.635 & 0.251 \\
& AB & \bf 0.256 & \bf 0.529 & \bf 0.761 & \bf 1.016 & \bf 0.551 &\bf  0.204 \\
\midrule
\multirow{2}{*}{table 3}
& MiDaS & \bf  0.656 & \bf  0.918 & \bf  0.985 &  0.833 & \bf  0.173 & \bf  0.086 \\
& AB & 0.615 & 0.896 & 0.972 & \bf 0.786 & 0.180 & 0.095 \\
\midrule
\multirow{2}{*}{table 4}
& MiDaS & \bf 0.704 &  \bf 0.956 & \bf 0.989 &  0.840 & 0.137 & \bf 0.077 \\
& AB & 0.694 & 0.951 & \bf 0.989 & \bf  0.804 & \bf 0.135 & 0.079 \\
\midrule
\multirow{2}{*}{table 5}
& MiDaS & 0.555 &  0.855 &  \bf 0.963 &  \bf 1.033 & 0.173 & 0.106 \\
& AB & \bf 0.571 & \bf 0.858 & 0.958 & 1.052 & \bf 0.171 & \bf 0.105 \\
\midrule
\multirow{2}{*}{table 6}
& MiDaS & 0.529 & \bf 0.826 &  \bf 0.955 &  1.012 & 0.220 & \bf 0.114 \\
& AB & \bf 0.532 & 0.816 & 0.938 & \bf 0.992 & \bf 0.217 & 0.116 \\
\midrule
\multirow{2}{*}{table 7}
& MiDaS &  \bf 0.590 & \bf  0.883 & \bf  0.973 &  \bf 0.991 & \bf  0.145 & \bf  0.098 \\
& AB & 0.513 & 0.841 & 0.955 & 1.069 & 0.147 & 0.113 \\
\midrule
\multirow{2}{*}{table 8}
& MiDaS & \bf  0.231 & \bf  0.561 & 0.873 & \bf  1.394 & 0.217 &  \bf 0.178 \\
& AB & 0.213 & 0.553 & \bf 0.878 & 1.508 & \bf 0.209 & 0.179 \\
\midrule
\multirow{2}{*}{Average}
& MiDaS & \bf  0.523 & 0.790 & 0.920 &  1.015 & 0.228 & 0.123 \\
& AB & 0.515 & \bf 0.800 & \bf 0.930 & \bf 0.998 & \bf 0.217 & \bf 0.121 \\
\bottomrule
\end{tabular}}
\end{table}

As  downstream applications such as AR and path planning may require depth (which the proposed AB-VINS is able to provide),
we evaluate the accuracy of the depth on the AR Table dataset.
Table~\ref{tab:ar_table_depth} reports the results of evaluating the dense depth output of AB-VINS.
We compare the base accuracy of the MiDaS DPT Swin2 Tiny model (denoted as ``MiDaS'' in the table) to the accuracy of the depth after jointly estimating $a$, $b$, $\mathbf{c}$, and $\mathbf{d}$ in our local mapping optimization (denoted as ``AB'' in the table).
The scale and bias for MiDaS are estimated linearly using sparse depths from the local map (i.e., the same points as tracking to static map and depth map registration) using the linear system from the original paper~\cite{Ranftl2022TPAMI} for a fair comparison.
From Table~\ref{tab:ar_table_depth} it is clear that estimating the AB features results in slightly improved depth accuracy -- improving all but one of the metrics in the average case.
Additionally, the dense depth accuracy of both MiDaS and the AB features is amenable for many downstream applications, with a $\delta_1$ accuracy over 0.5 and RMSE around one meter.

\begin{figure*}
\centering
\includegraphics[width=0.49\textwidth,trim={.1cm 1.2cm 1.1cm 0.9cm}, clip]{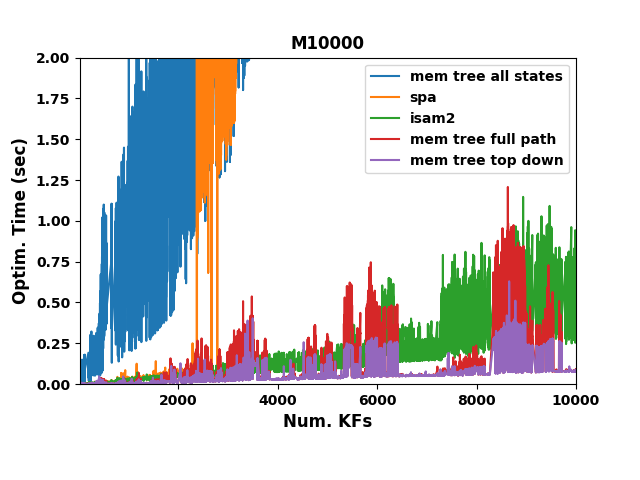}
\includegraphics[width=0.49\textwidth,trim={.1cm 1.2cm 1.1cm 0.9cm}, clip]{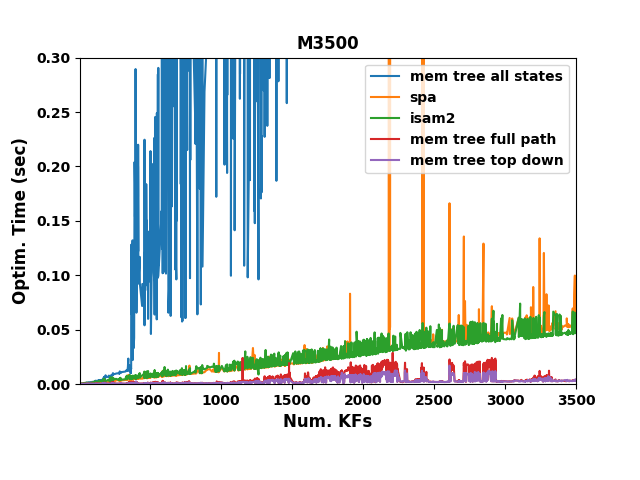}
\includegraphics[width=0.49\textwidth,trim={.1cm 1.2cm 1.1cm 0.cm}, clip]{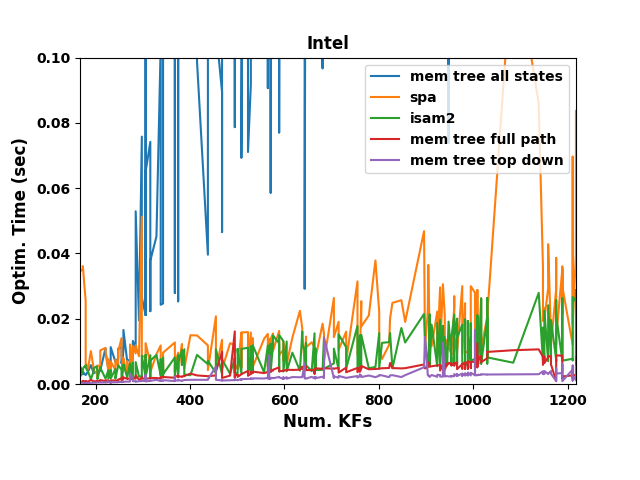}
\includegraphics[width=0.49\textwidth,trim={.1cm 1.2cm 1.1cm 0.cm}, clip]{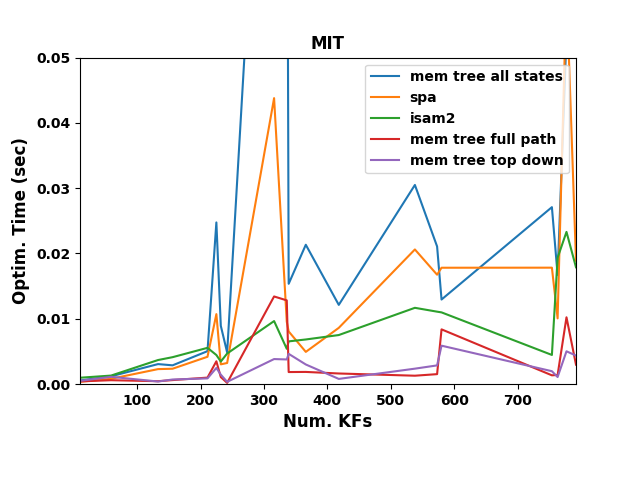}
\caption{Timing of different incremental pose graph optimization methods on the different 2D pose graph datasets.
Our memory-tree top-down method is by far the most efficient.}
\label{fig:tree_opt_timing}
\end{figure*}

\subsubsection{Accuracy of the Memory Tree}

To investigate the accuracy of the memory tree, we turn to the standard 2D pose graph datasets\footnote{The datasets are publicly available: \url{https://lucacarlone.mit.edu/datasets/}}.
We incrementally optimize the Manhattan World (M10000 and M3500) datasets as well as the Intel and MIT Killian Court datasets to simulate large-scale SLAM settings.
Optimization is only carried out when there is a new loop measurement.
For the memory tree, which performs 4-DoF pose graph optimization, the $z$ positions are all set to zero in order to utilize the 2D (3-DoF) datasets.
Since we only implement 4-DoF (3D position and yaw) pose graph optimization in the memory tree, the 3D (6-DoF) pose graph datasets (e.g., Sphere, Taurus, and Cube) are omitted.
We compare to the state-of-the-art incremental pose graph optimization method iSAM2~\cite{Kaess2011IJRR} implemented in GTSAM~\cite{gtsam}.
We also compare the three methods for optimizing the memory tree: 1) optimizing all states, 2) optimizing the full path between two loop nodes, and 3) the top-down optimization method. 
The resulting trajectories for the M10000 dataset can be viewed in Fig.~\ref{fig:tree_opt_traj_M10000}, the M3500 dataset in Fig.~\ref{fig:tree_opt_traj_M3500}, the Intel dataset in Fig.~\ref{fig:tree_opt_traj_INTEL}, and the MIT dataset in Fig.~\ref{fig:tree_opt_traj_MIT}.
Note that the memory tree trajectories are typically tilted compared to the iSAM2 trajectory due to the fact that the first frame can not be trivially fixed and has no prior.
Again, the global pose output can simply be taken relative to the first ever keyframe in order to match the output of standard methods with the first global pose fixed, but we left the trajectories tilted to show that we do not fix the first pose.
Qualitatively it can be seen that optimizing all states in the memory tree results in a trajectory similar to iSAM2, while the other two memory tree methods are slightly less accurate.
Of course in the Manhattan World datasets, a perfect solution would have all of the vertical parts of the trajectory at $90^\circ$ to the horizontal ones. 
However, the memory tree full path and top-down methods still produce reasonable results despite optimizing far fewer variables than is typical.
On the Intel dataset, there is almost no noticeable degradation in accuracy when using the memory tree full path or top-down method, which shows a case where these methods can nearly match the state-of-the-art solution despite being far more efficient.
For the MIT dataset, the performance of these two methods is noticeably worse than iSAM2 and the full batch memory tree solution.
While the accuracy is still reasonable for most of the trajectory (e.g., the smaller loops), we believe that the larger loops are less accurate due to the poor odometry accuracy on this dataset.
The memory tree full path and top-down methods require highly-accurate odometry in order to produce reasonable results, whereas on the MIT dataset the odemetry curves while traversing the long hallways and only optimizing a few states cannot correct all of the drift needed.

\subsection{Efficiency}

We now evaluate the efficiency of the proposed AB-VINS.

\begin{figure*}
\centering
\includegraphics[width=0.49\textwidth,trim={.1cm 1.2cm 1.1cm 0.9cm}, clip]{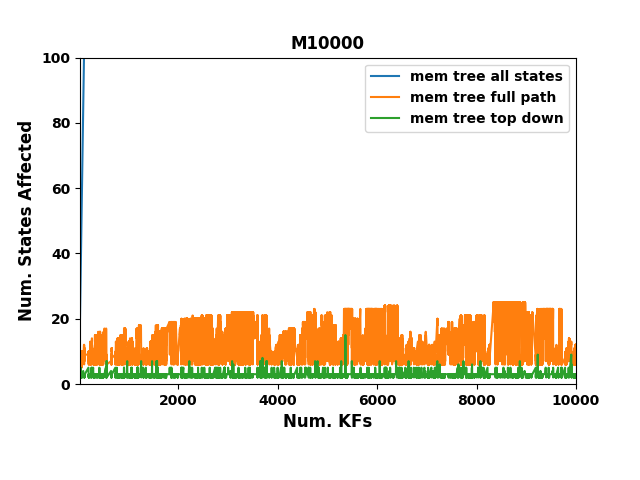}
\includegraphics[width=0.49\textwidth,trim={.1cm 1.2cm 1.1cm 0.9cm}, clip]{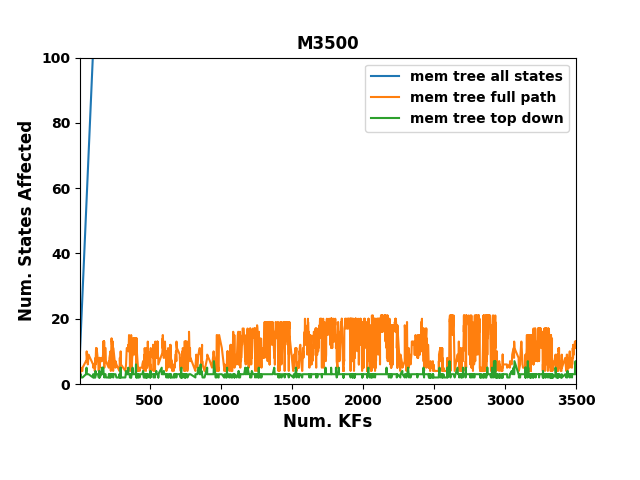}
\includegraphics[width=0.49\textwidth,trim={.1cm 1.2cm 1.1cm 0.cm}, clip]{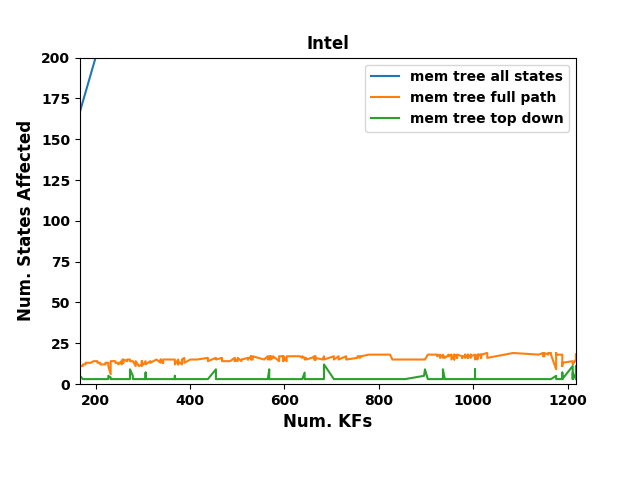}
\includegraphics[width=0.49\textwidth,trim={.1cm 1.2cm 1.1cm 0.cm}, clip]{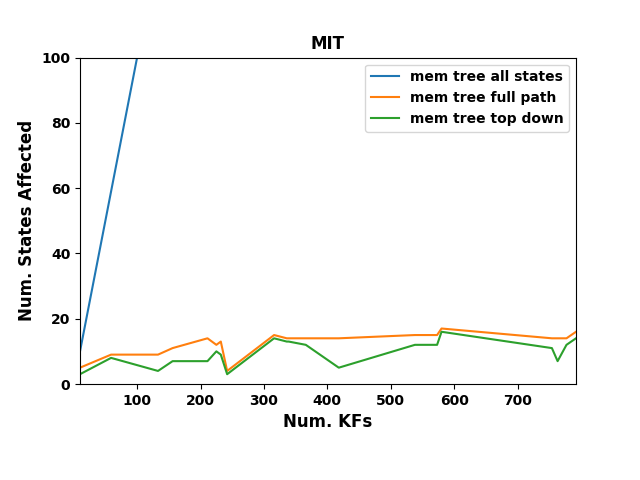}
\caption{The number of states affected by optimization over time on the different 2D pose graph datasets. 
The memory tree top-down method only affects a {\em constant} number of states on average.}
\label{fig:tree_opt_num_states}
\end{figure*}

\begin{figure}
\centering
\includegraphics[width=0.98\columnwidth,trim={.1cm 1.2cm 1.1cm 1.2cm}, clip]{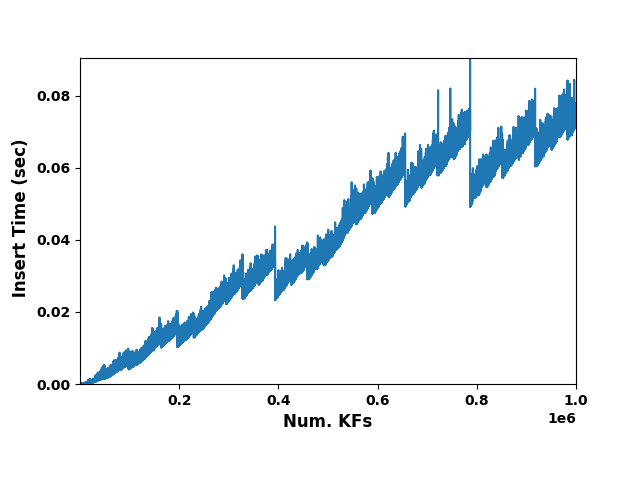}
\caption{Timing for inserting the newest node into the memory tree and balancing.}
\label{fig:tree_insert}
\end{figure}

\begin{figure}
\centering
\includegraphics[width=0.99\columnwidth,trim={.1cm 1.2cm 1.1cm 1.4cm}, clip]{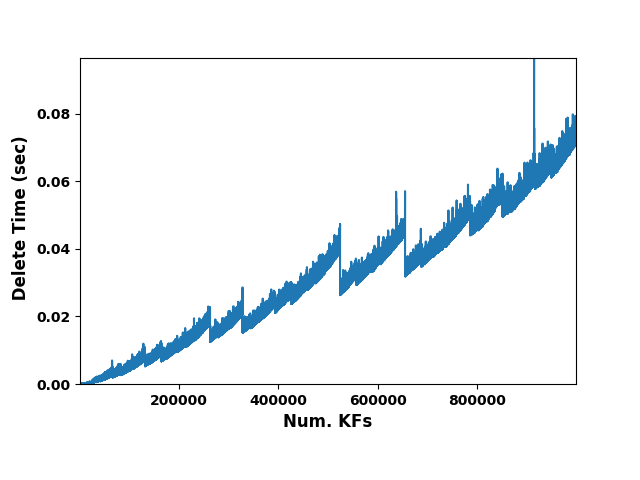}
\caption{Timing for deleting the most recent node in the memory tree and balancing.}
\label{fig:tree_delete}
\end{figure}

\begin{figure}
\centering
\includegraphics[width=0.99\columnwidth,trim={.1cm 1.2cm 1.1cm 1.1cm}, clip]{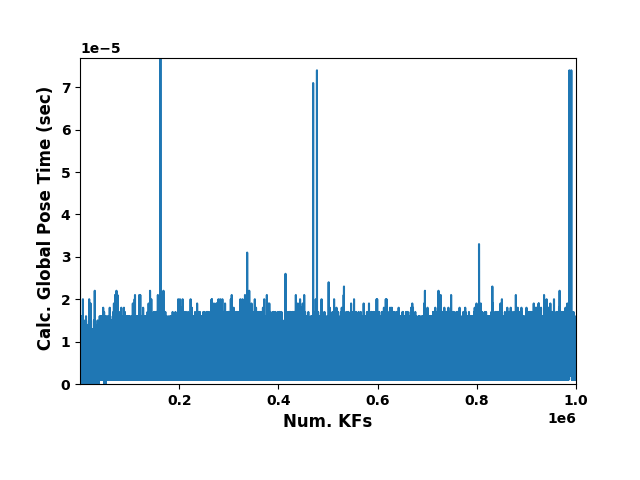}
\caption{Timing for calculating the global pose of the newest node in the memory tree.
Calculating the global pose never takes more than a handful of microseconds even with one million nodes despite being of logarithmic complexity.}
\label{fig:tree_calc}
\end{figure}

\begin{figure}
    \centering
\includegraphics[width=.99\columnwidth,trim={1.2cm 0cm 2.5cm 1cm}, clip]{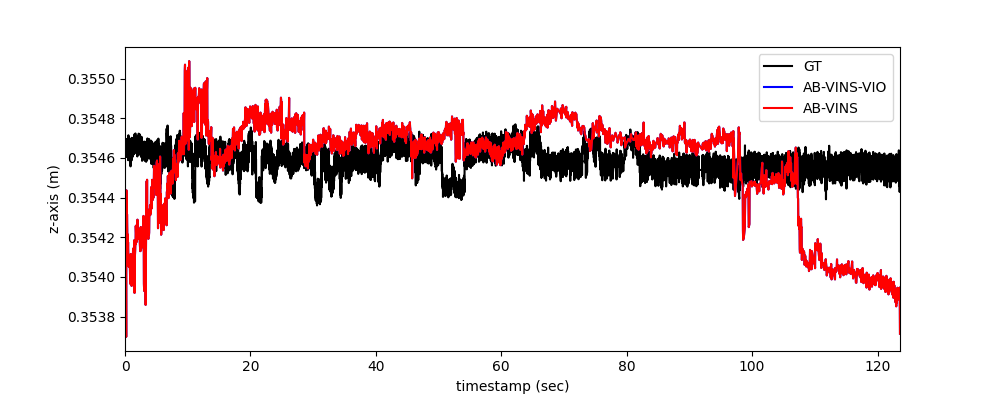}
    \includegraphics[width=.99\columnwidth,trim={1.2cm 0cm 2.5cm 1cm}, clip]{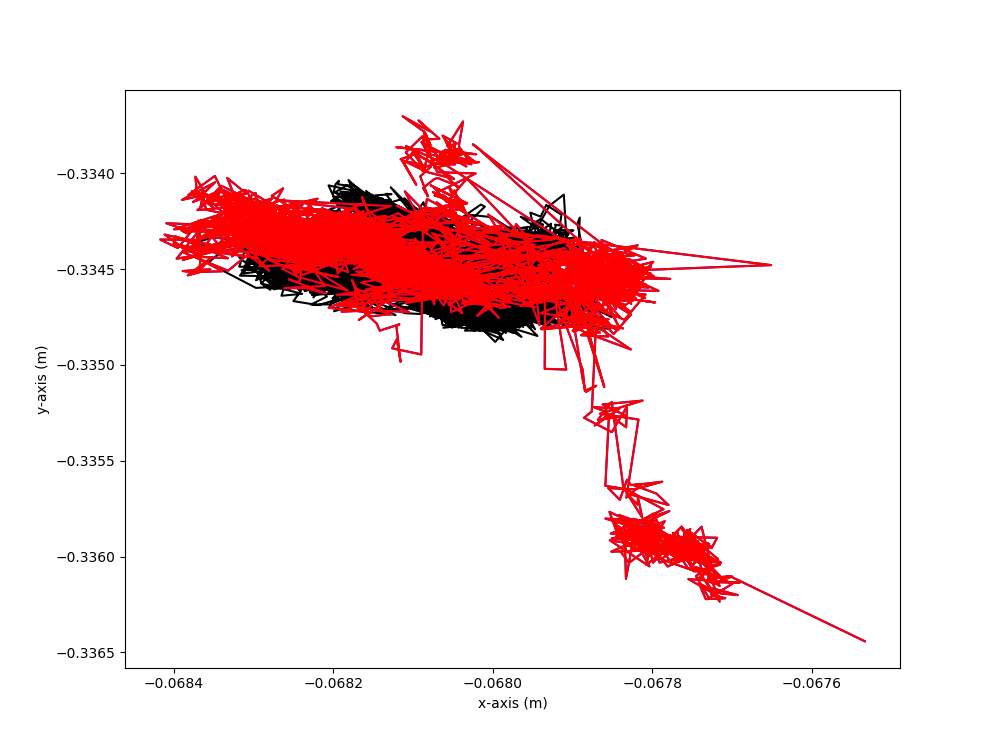}
    \caption{The trajectories from the \texttt{static} sequence of the Low/No Excitation dataset. The trajectories for VIO and AB-VINS are equivalent here and completely overlapping since there is only one keyframe. The trajectories look jagged only because of the extremely small scale of the plots.}
    \label{fig:static_traj}
\end{figure}

\begin{figure}
    \centering
\includegraphics[width=.99\columnwidth,trim={1.2cm 0cm 2.5cm 1cm}, clip]{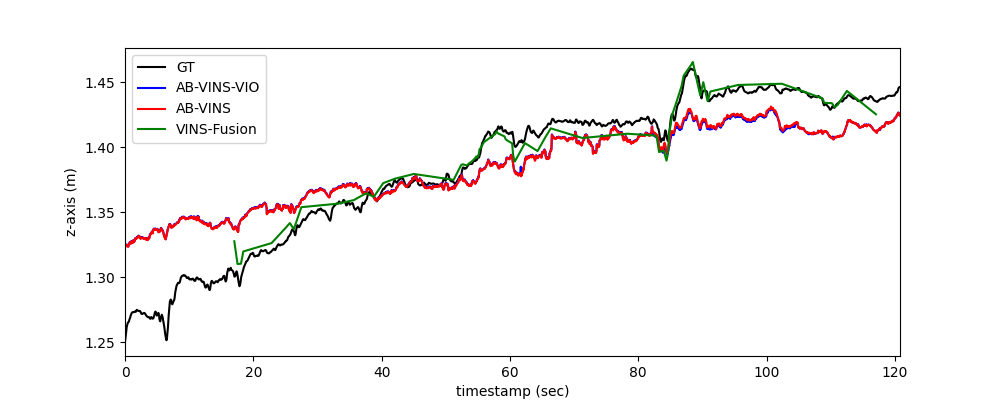}
    \includegraphics[width=.99\columnwidth,trim={1.2cm 0cm 2.5cm 1cm}, clip]{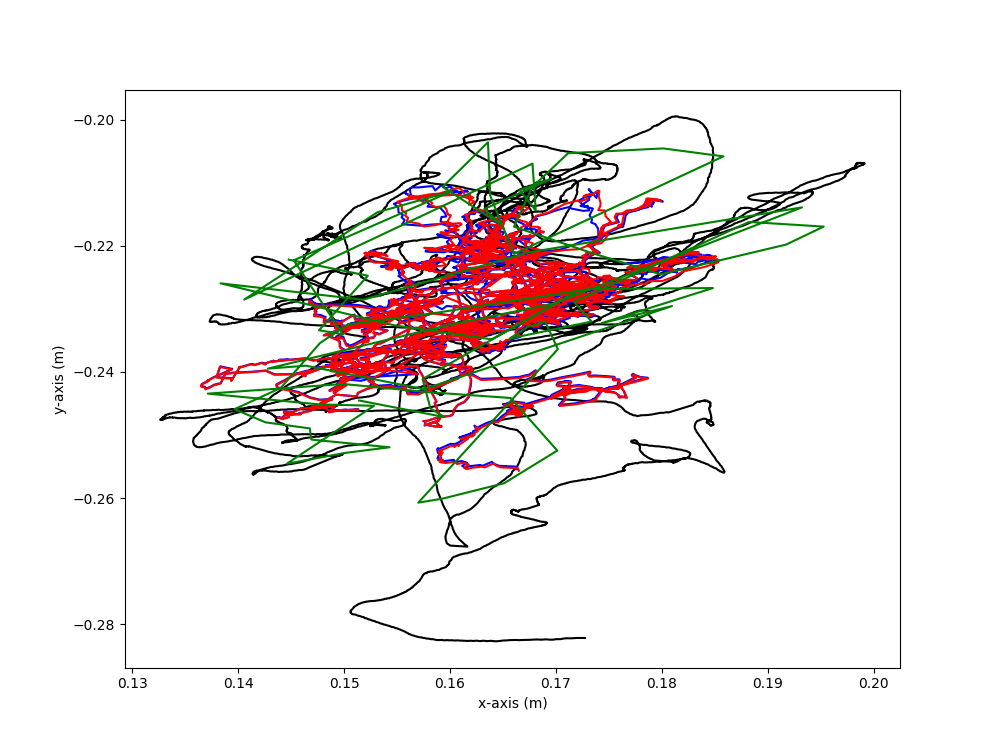}
    \caption{The trajectory plots from the \texttt{semi-static} sequence of the Low/No Excitation dataset. VINS-Fusion is surprisingly able to initialize on this sequence -- probably due to the slight $z$ motion induced by human error of trying to hold the sensors still.}
    \label{fig:semi_static_traj}
\end{figure}

\begin{figure}
    \centering
\includegraphics[width=.99\columnwidth,trim={1.2cm 0cm 2.5cm 1cm}, clip]{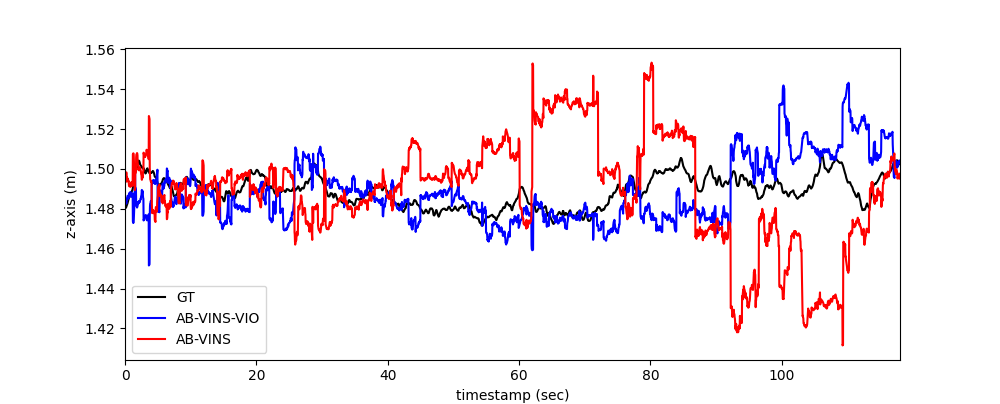}
    \includegraphics[width=.99\columnwidth,trim={1.2cm 0cm 2.5cm 1cm}, clip]{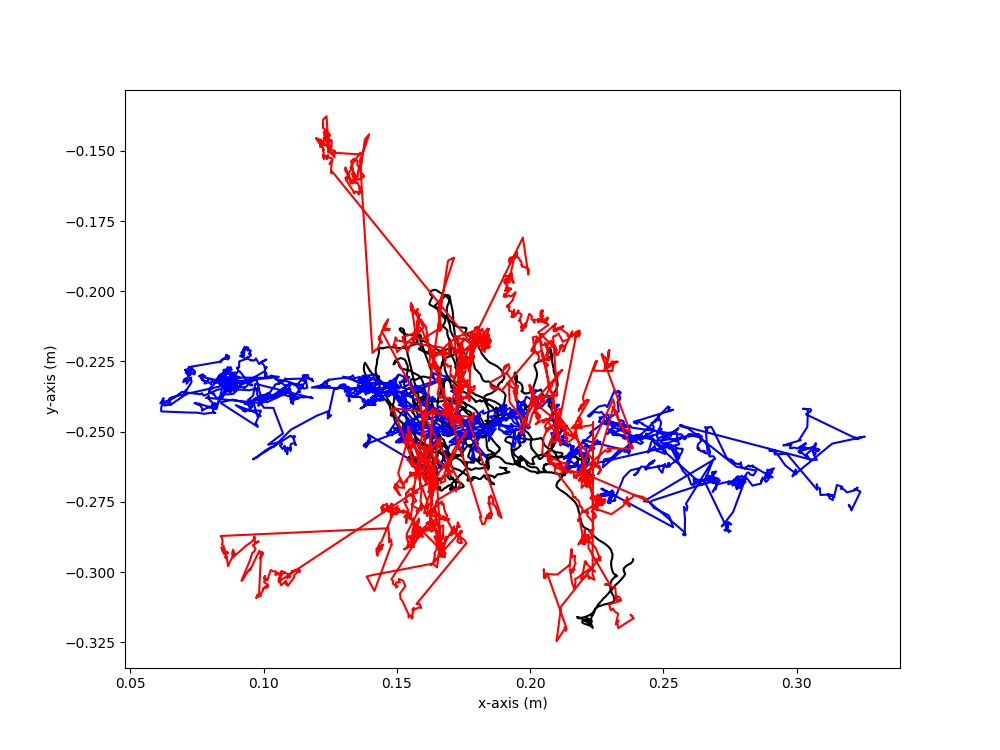}
    \caption{The trajectory plots from the \texttt{rotation} sequence of the Low/No Excitation dataset. Only AB-VINS is able to initialize on this challenging sequence.}
    \label{fig:rotation_traj}
\end{figure}

\begin{figure}
    \centering
\includegraphics[width=.99\columnwidth,trim={1.2cm 0cm 2.5cm 1cm}, clip]{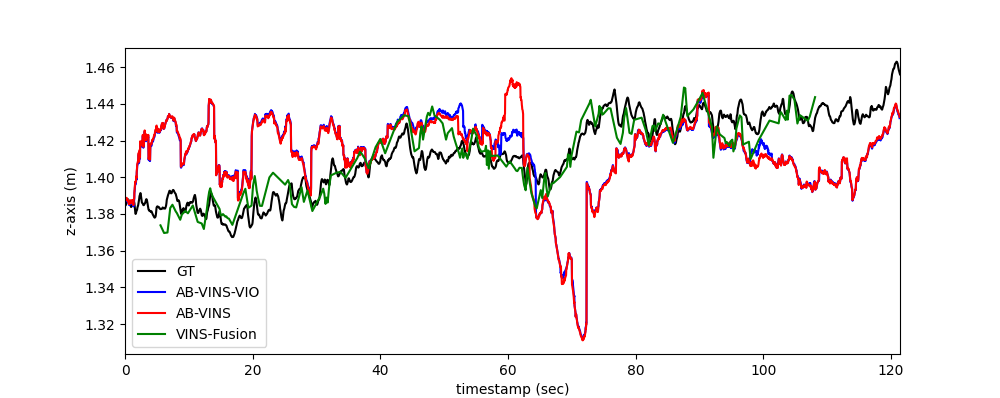}
    \includegraphics[width=.99\columnwidth,trim={1.2cm 0cm 2.5cm 1cm}, clip]{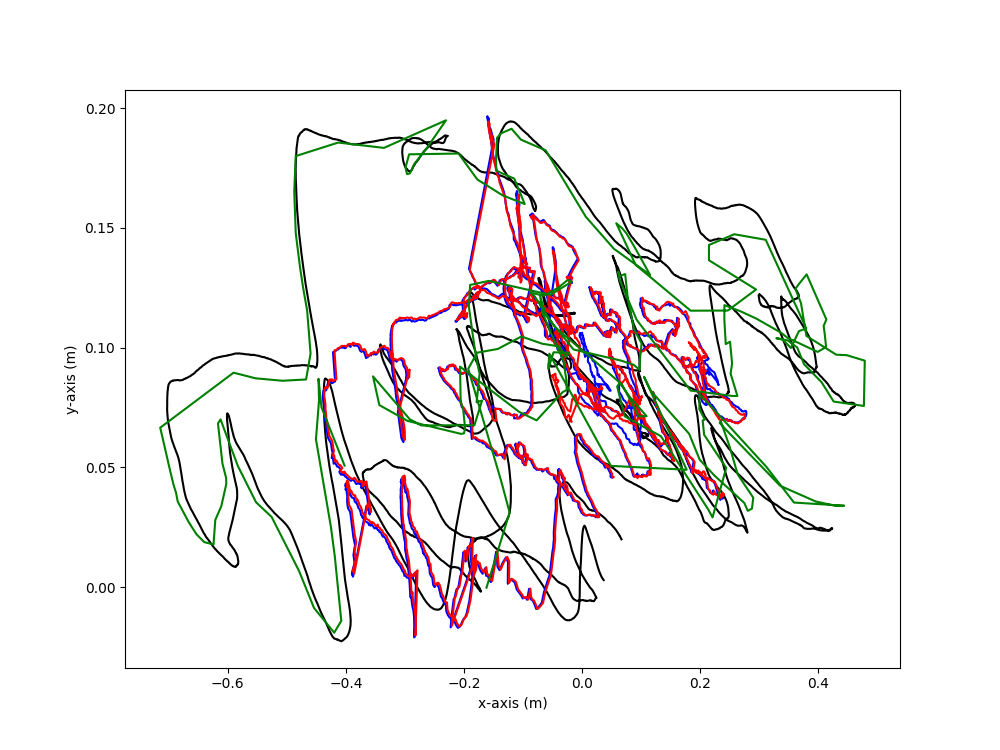}
    \caption{The trajectory plots from the \texttt{motion 1} sequence of the Low/No Excitation dataset. VINS-Fusion is the only fully hand-crafted system to initialize here.}
    \label{fig:motion_1_traj}
\end{figure}

\begin{figure}
    \centering
\includegraphics[width=.99\columnwidth,trim={1.2cm 0cm 2.5cm 1cm}, clip]{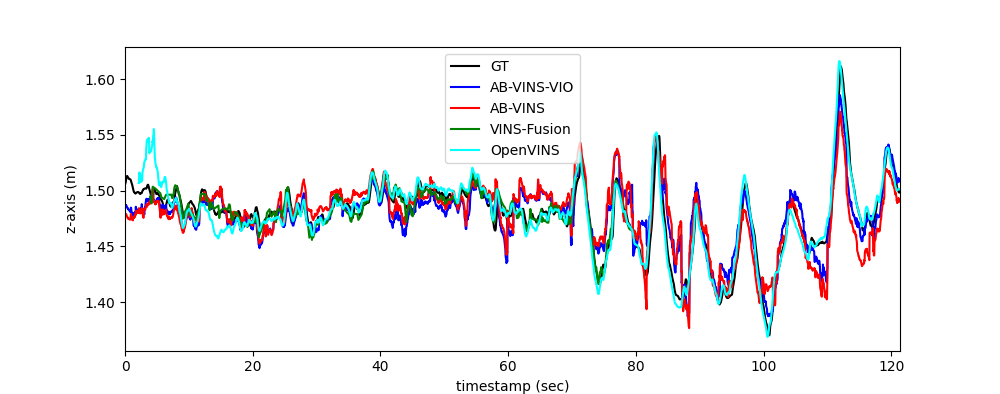}
    \includegraphics[width=.99\columnwidth,trim={1.2cm 0cm 2.5cm 1cm}, clip]{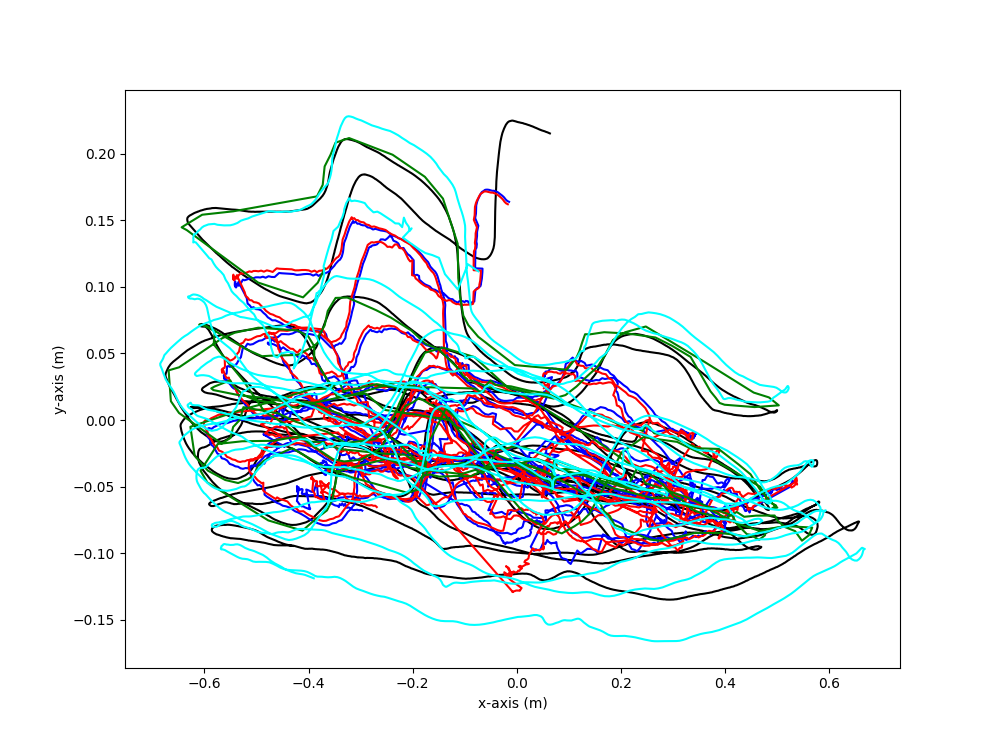}
    \caption{The trajectory plots from the \texttt{motion 2} sequence of the Low/No Excitation dataset. OpenVINS is able to now initialize with this amount of motion.}
    \label{fig:motion_2_traj}
\end{figure}

\begin{figure}
    \centering
\includegraphics[width=.99\columnwidth,trim={1.2cm 0cm 2.5cm 1cm}, clip]{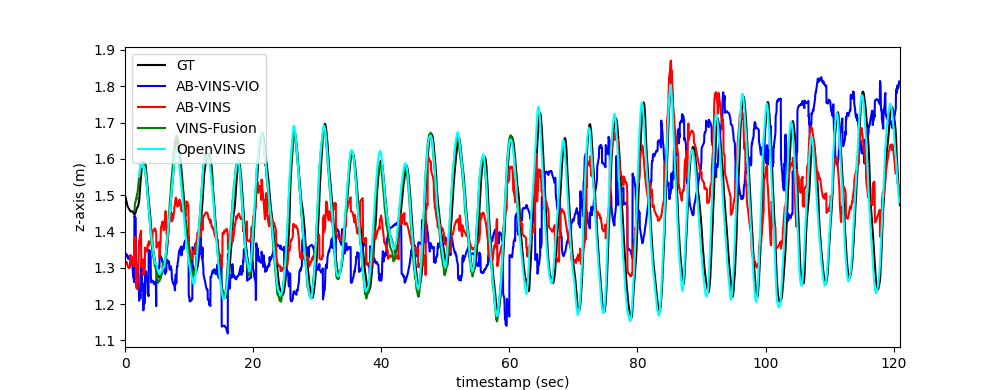}
    \includegraphics[width=.99\columnwidth,trim={1.2cm 0cm 2.5cm 1cm}, clip]{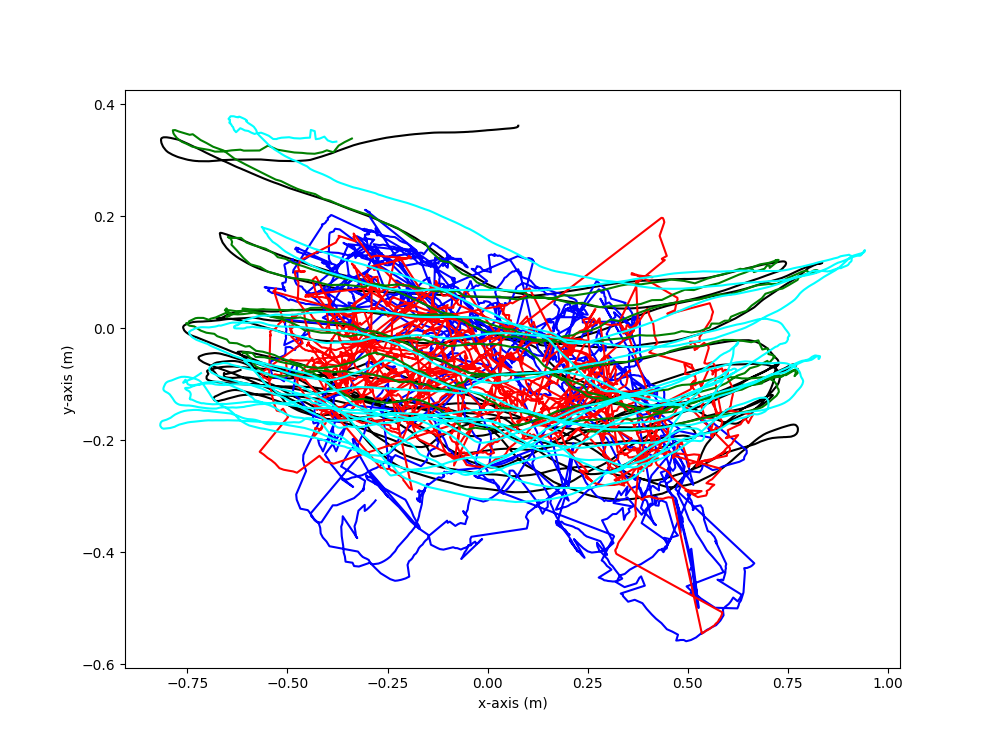}
    \caption{The trajectory plots from the \texttt{motion 3} sequence of the Low/No Excitation dataset. Despite there being more than a reasonable amount of motion for a typical AR/VR scenario, neither OKVIS or ORB-SLAM3 are able to initialize.}
    \label{fig:motion_3_traj}
\end{figure}

\subsubsection{Efficiency of the VIO Thread}

\begin{table} \centering
\caption{
Efficiency of operations in the VIO thread.
}
\label{tab:main_thread}
\resizebox{0.99\columnwidth}{!}{\begin{tabular}{@{}cc|cc@{}}
\toprule
\multicolumn{2}{c|}{\textbf{Frame Rate Operations}} & \multicolumn{2}{c}{\textbf{Keyframe Operations}} \\ 
\midrule
\textbf{Operation} & \textbf{Avg. Time (sec)} & \textbf{Operation} & \textbf{Avg. Time (sec)} \\
\midrule
Feature Tracking & $0.0033 \pm 0.0018$ & Mono Depth Inference & $0.0173 \pm 0.0026$ \\
Vision Only & $0.0032 \pm 0.0016$ & Depth Map Registration & $0.0048 \pm 0.0030$ \\
Inertial Only & $0.0001 \pm 0.0002$ & Local Mapping Opt. & $0.0701 \pm 0.0345$ \\
\midrule
Total Avg. & $0.0115 \pm 0.0233$ & & \\ \bottomrule \\
\end{tabular}}
\end{table}

\begin{table} \centering
\caption{
Average timing on the AR Table 4 sequence.
}
\label{tab:timing_comp}
\begin{tabular}{@{}cccc@{}}
\toprule
\textbf{Method}
 & \textbf{Avg. Time (sec)} & \textbf{Estimator} & \textbf{Dense Depth} \\ \midrule
VINS-Fusion~\cite{Qin2018TRO} & $0.0523 \pm 0.0260$ & Optimization & No \\
OpenVINS~\cite{Geneva2020ICRA} & $0.0123 \pm 0.0062$ & EKF & No \\
AB-VINS & $\mathbf{0.0115 \pm 0.0233}$ & Optimization & \bf Yes \\
\bottomrule \\
\end{tabular}\end{table}

Table~\ref{tab:main_thread} reports the efficiency of different operations of the main VIO thread,
namely frame rate operations (feature tracking, vision-only, and inertial-only optimization) as well as keyframe operations (mono depth inference, depth map registration, and local mapping optimization).
The total average time per frame is also reported as 11.5ms, showing that the main VIO thread can easily keep up with real-time requirements.
Note that the total average time per frame reported in Table~\ref{tab:main_thread} includes the keyframe operations (monocular depth inference, depth map registration, and local mapping optimization) in the average.
Table~\ref{tab:timing_comp} shows the comparison against the state-of-the-art VIO systems.
Only methods that use a single VIO thread are reported in the table, which is why ORB-SLAM3 and OKVIS are excluded.
VINS-Fusion's VIO is run in single-thread mode here to obtain the comparison. 
AB-VINS is much more efficient than a state-of-the-art optimization-based VINS-Fusion, and even surpasses the efficiency of a state-of-the-art filter-based OpenVINS on average while also providing dense depth.
Note that launching a separate thread for local mapping as in ORB-SLAM3 would further improve the efficiency of the main thread, since only feature tracking, vision only, and inertial only would be performed on it.

\subsubsection{Efficiency of the Memory Tree}

\begin{table*} \centering
\caption{
Position ATE (m) on the Low/No Excitation dataset.
}
\label{tab:low_no_ate}
\begin{tabular}{@{}cccccccccc@{}}
\toprule
\textbf{Type}  & \textbf{Algorithm}
 & \textbf{static} & \textbf{semi-static} & \textbf{rotation} & \textbf{motion 1} & \textbf{motion 2} & \textbf{motion 3} & \textbf{Average} \\ \midrule
\multirow{3}{*}{VIO}
& OKVIS~\cite{Leutenegger2014IJRR} & - & - & - & - & - & - & -${}^\dag$ \\
& OpenVINS~\cite{Geneva2020ICRA} & - & - & - & - & 0.044 & 0.109 & 0.077${}^*$ \\
& AB-VINS VIO & 0.001 & 0.029 & 0.057 & 0.123 & 0.099 & 0.357 & \underline{0.111} \\
\midrule
\multirow{3}{*}{SLAM}
& VINS-Fusion~\cite{Qin2018TRO} & - & 0.008 & - & 0.022 & 0.040 & 0.069 & 0.035${}^*$\\
& ORB-SLAM3~\cite{Campos2021TRO} & - & - & - & - & - & - & -${}^\dag$ \\
& AB-VINS & 0.001 & 0.029 & 0.051 & 0.124 & 0.104 & 0.287 & \bf{0.099} \\
\bottomrule \\
\multicolumn{8}{l}{${}^\dag$\footnotesize{Failed to initialize on all sequences so average could not be obtained.}} \\
\multicolumn{8}{l}{${}^*$\footnotesize{Failed to initialize on at least one sequence so not considered to be bolded or underlined.}}
\end{tabular}\end{table*}

\begin{figure*}
\centering
\includegraphics[width=0.31\textwidth,trim={0 0 1.5cm .5cm},clip]{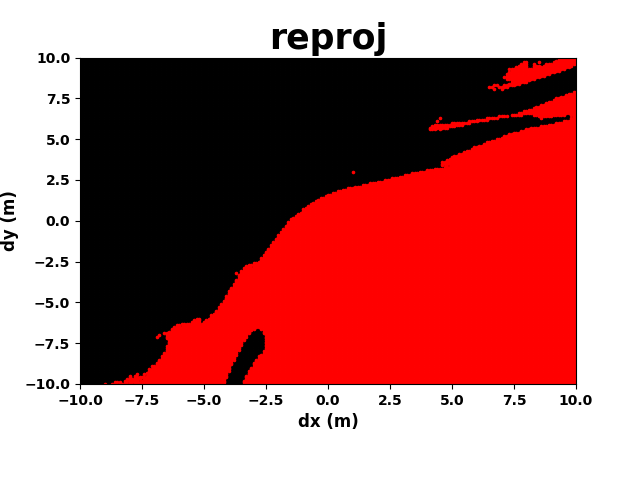}
\includegraphics[width=0.31\textwidth,trim={0 0 1.5cm .5cm},clip]{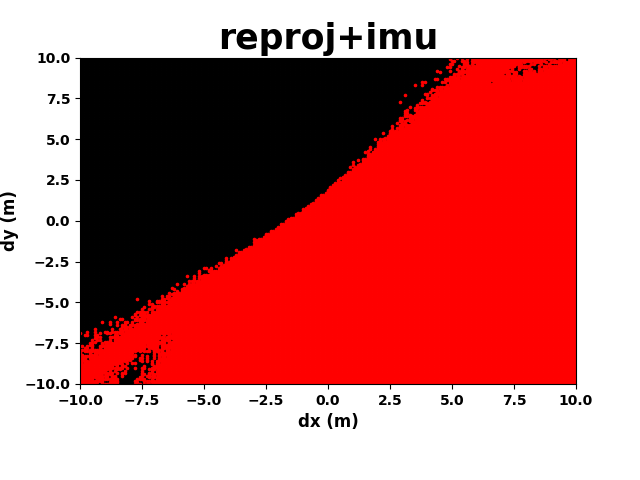}
\includegraphics[width=0.31\textwidth,trim={0 0 1.5cm .5cm},clip]{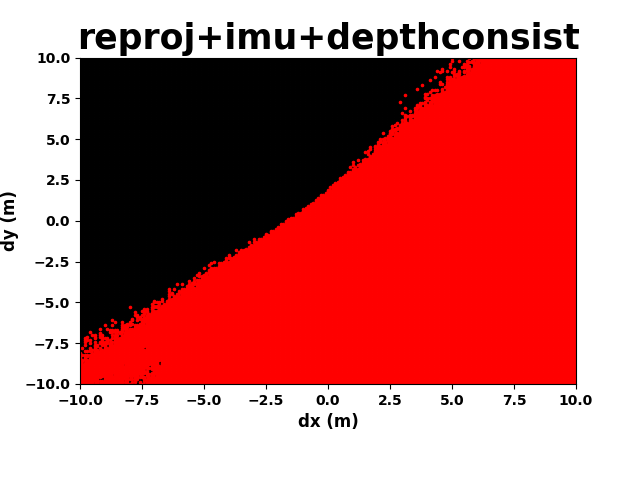}
\caption{The basin of attraction for \textbf{left}: visual BA, \textbf{center}: VI-BA, \textbf{right}: AB-VINS VI-BA with depth consistency. Converged points are in red, and diverged in black. 
It can be observed that the AB-VINS VI-BA converges the most often, making it the most robust optimization of the three.
}
\label{fig:basin}
\end{figure*}

\begin{figure}
\centering
\includegraphics[width=0.99\columnwidth,trim={0.5cm 1.1cm 1.5cm 1.4cm},clip]{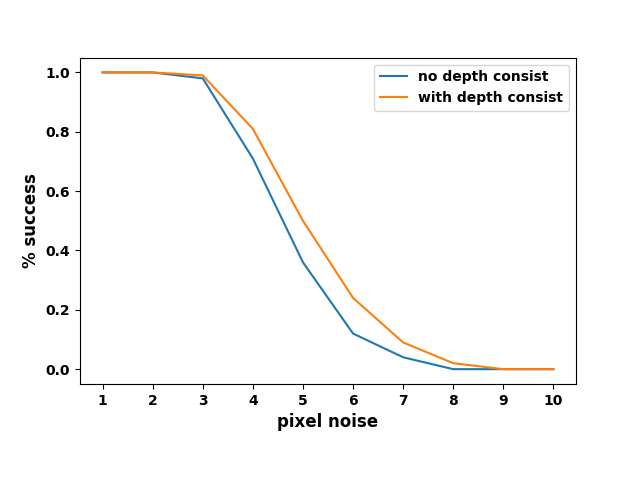}
\caption{Comparing the robustness to pixel noise for our VI-BA with depth consistency and a standard VI-BA. 
}
\label{fig:sim_pix_noise}
\end{figure}

\begin{figure*} 
\centering
\includegraphics[height=5.3cm,trim={.5cm 1cm 0cm .5cm},clip]{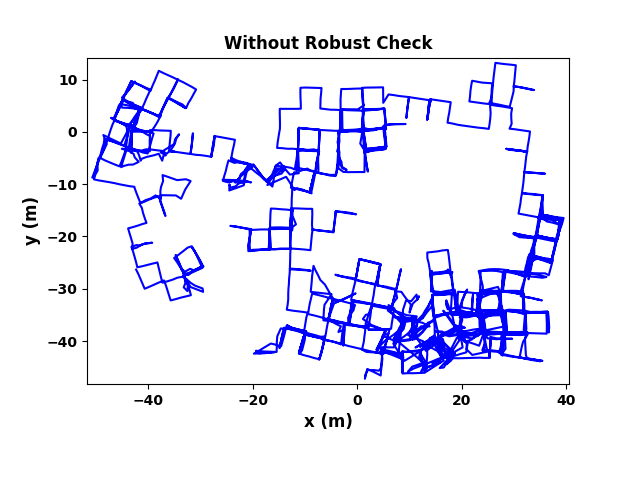}
\includegraphics[height=5.3cm,trim={0cm 1cm 1.9cm .5cm},clip]{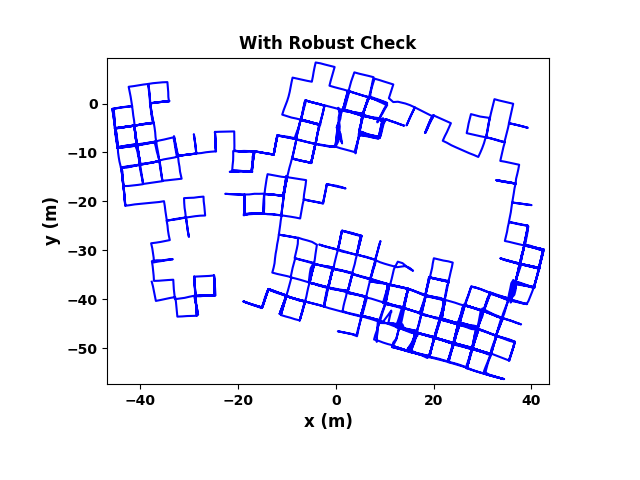}
\caption{The results of adding simulated bad loop measurements to the optimization. 
\textbf{Left}: the result without our robust $\chi^2$ check. 
\textbf{Right}: the result with our robust check.
}
\label{fig:tree_opt_robustness}
\end{figure*}

Where the full path and top-down memory tree methods lack in accuracy, they make up for in efficiency.
Timing results can be viewed in Fig.~\ref{fig:tree_opt_timing} for incrementally optimizing on the 2D pose graph datasets.
Here we also compare to SPA~\cite{Konolige2010IROS} for reference, which is reimplemented with Ceres solver using automatic differentiation for a fair comparison to our memory tree.
SPA was omitted in the accuracy evaluation since the SPA trajectories were very similar to the iSAM2 ones.
Note that SPA and iSAM2 have a slight advantage here since the implementation we use is purely 2D (3-DoF) instead of 4-DoF like ours.
Optimizing all states of the memory tree is very slow -- typically slower than SPA -- due to the extremely large and dense Hessian structure.
Both the full path and top-down methods are more efficient than SPA and iSAM2.
Besides the timing, it is also interesting to look at the state size of the optimization.
We claim that the top-down optimization solves pose graph SLAM while only adjusting a {\em constant} number of variables, and this is proven experimentally in Fig.~\ref{fig:tree_opt_num_states}.

We also provide timing for inserting a node into the memory tree -- the newest node as is done within AB-VINS -- and balancing.
This result can be viewed in Fig.~\ref{fig:tree_insert} for up to one million nodes.
Note that nearly all of the computation burden for memory tree insertion is because of the balancing, which is logarithmic with AVL trees but with a high constant overhead due to the cost of rotating the nodes on the way out of the recursive calls to the insert function.
The timing for deleting the most recent node from the memory tree can be seen in Fig.~\ref{fig:tree_delete}, where again most of the computational cost comes from balancing.
Finally, the cost of calculating the global pose for the most recent node in the memory tree can be seen in Fig.~\ref{fig:tree_calc}.
Even though calculating the global pose is logarithmic just like the complexity of balancing, the cost is far lower due to a much smaller number of floating point operations being performed, and only takes a few {\em microseconds} even with one million nodes in the tree.
This validates our design choice for the memory tree, since the global pose, which is needed for downstream applications, can be calculated very efficiently -- not taking much longer than having the pose defined in the global frame in the first place.

\subsection{Robustness} \label{sec:robustness}

We have also carefully validated the robustness of the proposed AB-VINS in different challenging scenarios.

\subsubsection{Robustness to Degenerate Motions}

It is known that completely hand-crafted monocular VINS  require parallax from sufficient motion to estimate the structure and poses.
On the other hand, since AB-VINS estimates the structure using monocular depth, it can work even with absolutely {\em no motion at all}.
To showcase this capability, we collected the Low/No Excitation dataset.
The dataset has six different motion profiles.
The first is \texttt{static}, where the device sits completely still on a table.
The second is called \texttt{semi-static}, where the device is held as still as possible in the hand.
After that, there is a sequence of rotation-only motion called \texttt{rotation}.
Finally, there are three levels of motion ranging from 1 to 3.
For \texttt{motion 1}, the device moves very slowly from side to side, and by \texttt{motion 3} the side-to-side motion is more aggressive and includes up-and-down motion as well as rotations.
All of the sequences are confined to a small area mostly facing the same thing in order to simulate a typical AR/VR scenario or low-motion robot activity (e.g., an autonomous vehicle sitting still at a traffic light or a drone hovering).
The same sensor rig as the AR Table dataset is used.
The results are reported in Table~\ref{tab:low_no_ate}.
Note that only the position error is reported here since the orientation error is impossible to calculate for the static sequences, and SE(3) alignment is used instead of the typical 4-DoF alignment since it is impossible to estimate the transform between the motion capture frame and the gravity-aligned frame with no motion in the trajectory.
It can be seen in Table~\ref{tab:low_no_ate} that AB-VINS (both in the VIO and SLAM setting) is the only system that can successfully initialize and estimate the poses in all sequences.
A system is determined to have failed in this experiment if there was no pose output (e.g., it did not attempt to initialize) or if the position error is over 10m.
OKVIS and ORB-SLAM3 fail to initialize in any of the sequences, while OpenVINS is successful in the final two, and VINS-Fusion is successful in all but the first completely static sequence and the rotation-only sequence.
The performance of VINS-Fusion is surprisingly very robust to low excitation for a completely hand-crafted system, but nevertheless AB-VINS can work in all of the scenarios.

We also provide visualizations of each trajectory in the dataset and the estimated trajectories for each system which successfully initialized.
The plots are shown in Figs.~\ref{fig:static_traj}, \ref{fig:semi_static_traj}, \ref{fig:rotation_traj}, \ref{fig:motion_1_traj}, \ref{fig:motion_2_traj}, and \ref{fig:motion_3_traj} for the \texttt{static}, \texttt{semi-static}, \texttt{rotation}, \texttt{motion 1}, \texttt{motion 2}, and \texttt{motion 3} sequences, respectively.
For the \texttt{static} sequence in Fig.~\ref{fig:static_traj}, AB-VINS is the only system which is able to initialize because of the lack of any motion at all.
The scale of the plot is very small, which leads to a jagged appearance of the trajectories even though they are smooth from a distance.
On the \texttt{semi-static} sequence in Fig.~\ref{fig:semi_static_traj}, VINS-Fusion is surprisingly able to initialize.
However, it is clear from the $z$-axis time-series plot that it takes longer than AB-VINS to start up, since it requires at least a small amount of motion -- which is present in a small amount in the $z$ direction due to human error in trying to hold the device still.
For the \texttt{rotation} sequence in Fig.~\ref{fig:rotation_traj}, VINS-Fusion fails to initialize again.
This is because the operator was able to hold the device more still in terms of translation than for the \texttt{semi-static} sequence, and rotation-only does not supply enough parallax for VINS-Fusion to triangulate features.
On the \texttt{motion 1} sequence in Fig.~\ref{fig:motion_1_traj}, VINS-Fusion begins to noticeably produce a better trajectory than AB-VINS, since sufficient parallax is now present.
However, it again takes longer to initialize -- delaying the startup of any downstream application.
For the \texttt{motion 2} and \texttt{motion 3} sequences in Figs.~\ref{fig:motion_2_traj} and~\ref{fig:motion_3_traj}, OpenVINS is now able to initialize and produce reasonable results.
However, again, the completely hand-crafted systems exhibit delayed initialization, while AB-VINS is able to start from the first frame.
Despite there being more motion in the \texttt{motion 3} sequence than a typical AR/VR scenario, OKVIS and ORB-SLAM3 fail to initialize.

\subsubsection{Robustness of Local Mapping Optimization}

\begin{table} \centering
\caption{
Optimization with 40,401 different initials.
}
\label{tab:basin}
\begin{tabular}{@{}ccc@{}}
\toprule
\textbf{Optimization}
 & \textbf{\# Success} & \textbf{\% Success} \\ \midrule
V-BA & 19177 & 47.5 \\
VI-BA & \underline{24338} & \underline{60.2} \\
VI-BA + depth consist. (Ours) & \bf{24592} & \bf{60.9} \\
\bottomrule \\
\end{tabular}\end{table}

To investigate the robustness of our local mapping optimization, we turn to simulation.
Using the \texttt{table 1} sequence of the AR Table dataset, we fit a BSpline to the trajectory.
The derivatives of the BSpline are used to simulate IMU measurements, and 3D landmarks are generated randomly and projected to simulate feature track and scale-less inverse depth measurements.
Camera measurements are generated at 1Hz to simulate keyframes, and two keyframes along with the measurements are fed to our local mapping optimization.

We investigate the basin of attraction for the optimization with 1) only visual measurements (V-BA), 2) visual and inertial measurements (VI-BA), and 3) our full VI-BA with depth consistency.
All feature parameters (3D landmarks and AB features) are fixed to the ground truth for this experiment.
The initial guess of the second IMU position is perturbed over a $10 \times 10$ meter grid in increments of 0.1m in order to investigate the basin of attraction.
The visual results are shown in Fig.~\ref{fig:basin}, where it can be observed that the VI-BA is more robust than the V-BA, and our VI-BA with depth consistency is the most robust of the three.
The number of successful runs is also reported in Table~\ref{tab:basin}.
Success (the attraction set) is determined here if the estimated IMU position converged within 1cm of the ground truth. 

We also investigate the robustness of local mapping to pixel noise.
We run the simulation with pixel noise varying from 1 to 10 -- averaged over 100 different random seeds per pixel noise.
We compare our full VI-BA with depth consistency to the standard VI-BA.
The results are shown in Fig.~\ref{fig:sim_pix_noise}, where it can be seen that our VI-BA is more robust than the standard one.
Note that the standard 1 pixel noise covariance value was provided to the estimators here, so the increased noise is unknown to the system.

\subsubsection{Robustness to IMU Saturation}
\begin{table} \centering
\caption{
Simulated IMU saturation over 1,000 trials.
}
\label{tab:imu_sat}
\begin{tabular}{@{}ccc@{}}
\toprule
\textbf{Preintegration Method}
 & \textbf{\# Success} & \textbf{\% Success} \\ \midrule
Standard & 8 & 0.8 \\
Ours (robust) & \bf{953} & \bf{95.3} \\
\bottomrule \\
\end{tabular}\end{table}

IMU saturation can occur during sharp motions or the sensors impacting on an object.
While most VINS algorithms are not robust to IMU saturation, AB-VINS is.
To show this, we ran the same simulation as in the previous section, but added saturated IMU measurements to a random axis for 1\% of the simulated gyroscope and accelerometer measurements.
We use the full VI-BA with depth consistency cost in this experiment.
The simulation is run with 1,000 different random seeds both with our robust method for IMU saturation (inflating the noise) and without.
It can be seen in Table~\ref{tab:imu_sat} that our robust noise inflation method is successful nearly all of the time, while the standard method fails almost all of the time.

\subsubsection{Robustness to Bad Loops}

Bad loop closure detection (e.g., from incorrect place recognition) can be catastrophic for SLAM.
Our robust $\chi^2$ check for loop measurements makes AB-VINS robust to incorrect loops.
To showcase this, using the M3500 dataset, we simulate bad loop measurements for 10\% of the loops by adding large random noise to the yaw and position portions of the measurements (standard normal noise in radians for yaw and larger noise for position with standard deviation ranging from 1 to 50 randomly).
The results can be seen in Fig.~\ref{fig:tree_opt_robustness}, where it can be observed that our robust $\chi^2$ check successfully rejects the majority of bad loops, while not using the check results in a visibly worse trajectory.
The memory tree top-down optimization method is used in both cases for this experiment. \section{Conclusions} \label{sec:conclusion}

We have presented a new visual-inertial SLAM system called AB-VINS, which utilizes three different deep networks.
We showed that it is possible, in AB-VINS, to only estimate $a$ and $b$ for monocular depth maps, as well as some other terms ($\mathbf{c}$ and $\mathbf{d}$) to correct the depth with multi-view information instead of each feature position separately.
The memory tree, a novel data structure to speed up pose graph optimization, has been introduced.
AB-VINS has been shown to have state-of-the-art efficiency and robustness while also providing dense depth.
AB-VINS is a different kind of VINS, 
one that heavily relies on deep learning as well as new and improved hand-crafted techniques, and prioritizes efficiency and robustness over accuracy.
With this new approach in the open, we hope that researchers can re-think how they perform VINS.

\section*{Acknowledgement}
We would like to thank Dr. Mingyang Li (Google and now Meta) for his constant support and discussion throughout the development of AB-VINS.
We would also like to thank our labmates Yuxiang Peng, Saimouli Katragadda, and Chuchu Chen for their help in collecting the low/no excitation dataset.
This work was partially supported by the University of Delaware (UD) College of Engineering, 
the NSF (IIS-1924897, SCH-2014264), and Google.

\bibliographystyle{ieeetr}
\bibliography{library,rpng}

\begin{IEEEbiography}[{\includegraphics[width=1in,height=1.25in,clip,keepaspectratio]{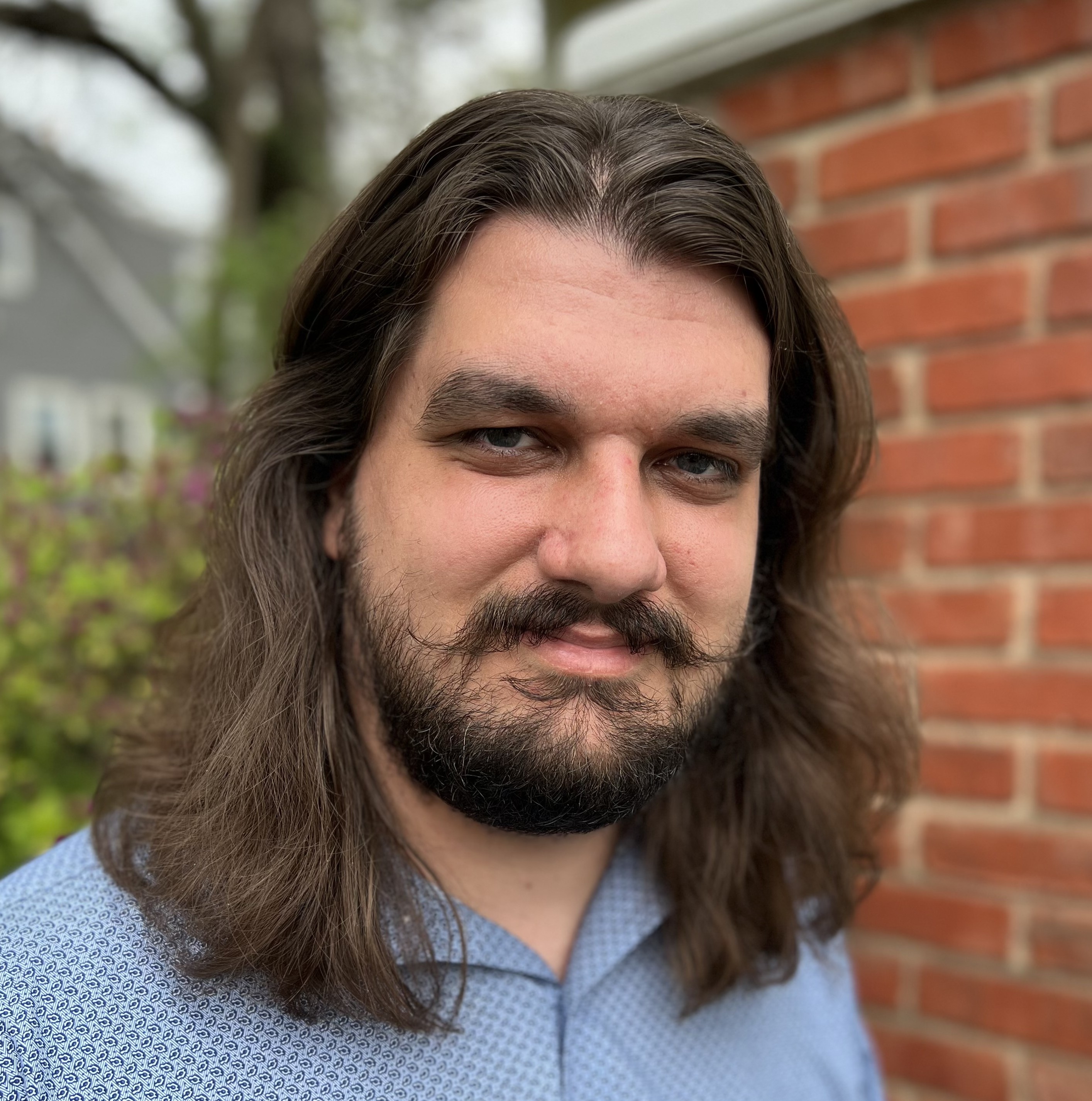}}]{Nathaniel Merrill}
received his B.S. in computer science from the University of Delaware in 2019.
He is currently finishing his Ph.D. in computer science at the University of Delaware.
His research interests include deep learning and VINS.

Nathaniel has received the First Place Intern Poster Award in computer science at NASA Goddard Space Flight Center in 2017, the AAUP-UD award in 2019, Robot Vision Award finalist at ICRA 2021, and Best Student Paper Award finalist at RSS 2023.
\end{IEEEbiography}

\begin{IEEEbiography}[{\includegraphics[width=1in,height=1.25in,clip,keepaspectratio]{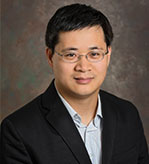}}]
{Guoquan Huang}(Senior Member, IEEE) received his BS in automation (electrical engineering) from  University of Science and Technology Beijing, China, in 2002, and MS and PhD in computer science from  University of Minnesota--Twin Cities, in 2009 and 2012, respectively. He currently is an Associate Professor of Mechanical Engineering (ME) and Computer and Information Sciences (CIS) at  University of Delaware (UD), where he is leading  Robot Perception and Navigation Group (RPNG).  From 2012 to 2014, he was a Postdoctoral Associate with MIT CSAIL (Marine Robotics). His research interests focus on state estimation and spatial computing for autonomous robots and mobile devices, including probabilistic sensing, estimation, localization, mapping, perception, locomotion and navigation. 
He has served as an Associate Editor for the IEEE Transactions on Robotics (T-RO) and IET Cyber-Systems and Robotics (CSR), as well as previously for the IEEE Robotics and Automation Letters (RA-L).
Dr. Huang has received 2015 UD Research Award (UDRF), 2015/2023 NASA DE Space Research Seed Award, 2016 NSF Research Initiation (CRII) Award, 
2018/2019/2022 Google Daydream/ARVR/AI Faculty Research Award, 
and 2023 Meta Reality Labs Faculty Research Award. 
He was the recipient of 2022 GNC Journal Best Paper Award, ICRA 2022 Best Paper Award (Navigation), and the Finalists of ICRA 2024 Best Paper Award (Robot Vision), RSS 2023 Best Student Paper Award, ICRA 2021 Best Paper Award (Robot Vision) and RSS 2009 Best Paper Award.

\end{IEEEbiography}\textbf{}

\end{document}